\documentclass[sigconf]{acmart}

\usepackage{csquotes}
\usepackage{url}
\usepackage{amsmath}
\usepackage{mathtools}
\usepackage{csquotes}
\usepackage{enumitem}
\usepackage{xspace}
\usepackage{algorithmic}
\usepackage{algorithm}
\usepackage{subcaption}
\usepackage{array}
\usepackage{mathtools} 
\usepackage{comment}
\usepackage[group-separator={,}]{siunitx}
\usepackage{bm}
\usepackage{placeins}
\usepackage{amsthm}
\usepackage{xcolor}
\usepackage{soul}
\usepackage{multirow}
\usepackage{makecell}
\usepackage{booktabs}
\usepackage{hyperref}

\AtBeginDocument{%
  \providecommand\BibTeX{{%
    \normalfont B\kern-0.5em{\scshape i\kern-0.25em b}\kern-0.8em\TeX}}}

\copyrightyear{2021}
\acmYear{2021}
\setcopyright{acmcopyright}\acmConference[KDD '21]{Proceedings of the 27th ACM SIGKDD Conference on Knowledge Discovery and Data Mining}{August 14--18, 2021}{Virtual Event, Singapore}
\acmBooktitle{Proceedings of the 27th ACM SIGKDD Conference on Knowledge Discovery and Data Mining (KDD '21), August 14--18, 2021, Virtual Event, Singapore}
\acmPrice{15.00}
\acmDOI{10.1145/3447548.3467222}
\acmISBN{978-1-4503-8332-5/21/08}

\begin{document}
\fancyhead{}

\title{Robust Learning by Self-Transition for Handling Noisy Labels}


\author{Hwanjun Song$^{1}$, Minseok Kim$^{2}$, Dongmin Park$^{2}$, Yooju Shin$^{2}$, Jae-Gil Lee$^{2,}$}
\authornote{Jae-Gil Lee is the corresponding author.}
\affiliation{
\institution{$^{1}$NAVER AI Lab $^{2}$Korea Advanced Institute of Science and Technology}
}
\email{hwanjun.song@navercorp.com, {minseokkim, dongminpark, yooju24, jaegil}@kaist.ac.kr}


\begin{abstract}
Real-world data inevitably contains noisy labels, which induce the poor generalization of deep neural networks. It is known that the network typically begins to rapidly memorize false-labeled samples after a certain point of training. Thus, to counter the label noise challenge, we propose a novel \emph{self-transitional} learning method called \textbf{\algname{}}, which automatically switches its learning phase at the transition point from \emph{seeding} to \emph{evolution}. In the seeding phase, the network is updated using all the samples to collect a seed of clean samples. Then, in the evolution phase, the network is updated using only the set of arguably clean samples, which precisely keeps expanding by the updated network. Thus, \algname{} effectively avoids the overfitting to false-labeled samples throughout the entire training period. Extensive experiments using five real-world or synthetic benchmark datasets demonstrate substantial improvements over state-of-the-art methods in terms of robustness and efficiency.
\end{abstract}



\newtheorem{assumption}[theorem]{Assumption}
\newcommand{\colorcomment}[3]{\xspace{\color{#2} [{#1}]:{#3}}\xspace}
\newcommand{\phaseI}{\textbf{Seeding}}
\newcommand{\phaseII}{\textbf{Evolution}}
\newcommand{\algname}{{MORPH}}
\newcommand{\default}{{Default}}
\newcommand{\coteaching}{{Co-teaching}}
\newcommand{\incv}{{INCV}}
\newcommand{\itlm}{{ITLM}}
\newcommand{\selfie}{{SELFIE}}
\renewcommand{\algorithmicrequire}{\textsc{Input:}}
\renewcommand{\algorithmicensure}{\textsc{Output:}}
\renewcommand{\algorithmiccomment}[1]{/*~#1~*/}
\newcommand{\INDSTATE}[1][1]{\STATE\hspace{#1\algorithmicindent}}
\DeclarePairedDelimiter{\ceil}{\lceil}{\rceil}

\newcolumntype{L}[1]{>{\raggedright\let\newline\\\arraybackslash\hspace{0pt}}m{#1}}
\newcolumntype{X}[1]{>{\centering\let\newline\\\arraybackslash\hspace{0pt}}p{#1}}

\maketitle

\sloppy

\section{Introduction}
\label{sec:intro}

In supervised learning for data analysis tasks, deep neural networks\,(DNNs) have become one of the most popular methods in that traditional machine learning is successfully superseded by recent deep learning in numerous applications\,\cite{tang2019msuru, kamani2020targeted}. However, their success is conditioned on the availability of massive data with carefully annotated human labels, which are expensive and time-consuming to obtain in practice. Some substitutable sources, such as Amazon's Mechanical Turk and surrounding tags of collected data, have been widely used to mitigate the high labeling cost, but they often yield samples with \emph{noisy labels} that may not be true\,\cite{song2020learning}. In addition, data labels can be extremely complex even for experts\,\cite{frenay2013classification} and adversarially manipulated by a label-flipping attack\,\cite{xiao2012adversarial}, thereby being vulnerable to label noise.

Modern DNNs are typically trained in an over-parameterized regime where the number of the parameters of a DNN far exceeds the size of the training data\,\cite{li2020gradient}. In principle, such DNNs have the capacity to overfit to any given set of labels, even though part of the lables are significantly corrupted. In the presence of noisy labels, DNNs easily overfit to the entire training data regardless of the ratio of noisy labels, eventually resulting in poor generalization on test data\,\cite{zhang2016understanding}. Thus, in this paper, we address an important issue of \enquote{learning from noisy labels.}

One of the most common approaches is \emph{sample selection}, which involves training a DNN for a possibly clean subset of noisy training data\,\cite{huang2019o2u, jiang2017mentornet, han2018co, shen2019learning, chen2019understanding, song2020prestopping}. Typically, in each training iteration, a certain number of \emph{small-loss} training samples are selected as clean ones and subsequently used to robustly update the DNN. This small-loss trick is satisfactorily justified by the \emph{memorization effect}\,\cite{arpit2017closer} that DNNs tend to first learn simple and generalized patterns and then gradually memorize all the patterns including irregular ones such as outliers and false-labeled samples. 

\begin{figure*}[ht!]
\begin{center}
\includegraphics[width=16cm]{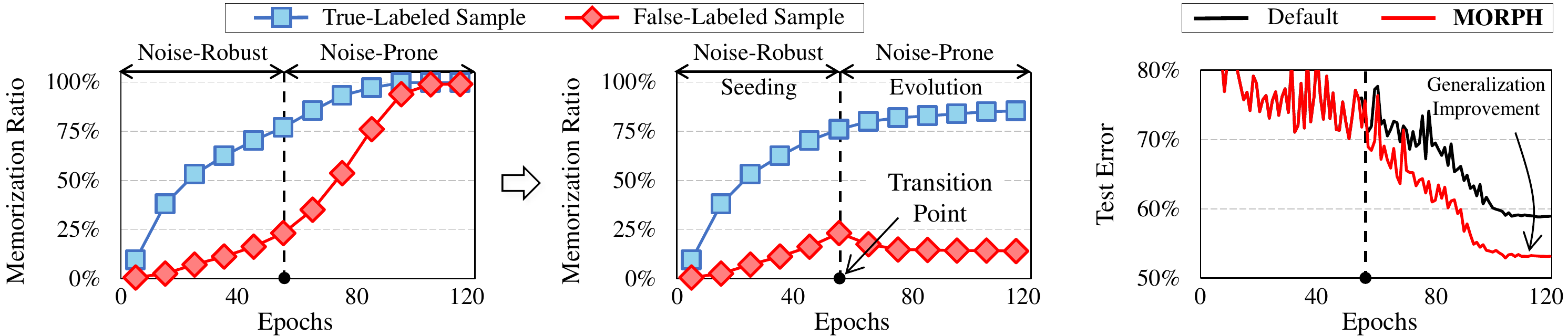}
\end{center}
\vspace*{-0.1cm}
\hspace*{1.95cm} {\small (a) Default.} \hspace*{4.25cm} {\small (b) MORPH.}  \hspace*{2.95cm} {\small (c) Test Error Convergence.}
\vspace*{-0.35cm}
\caption{Key idea of MORPH: (a) and (b) show the memorization ratio when training a WideResNet-16-8 on a subset of $\text{FOOD-101N}$\protect\footnotemark ~with the real-world noise of $18.4\%$, where the memorization ratio is the number of memorized\,(see Definition \ref{def:fully_learned_sample}) true- or false-labeled samples to the total number of true- or false-labeled training samples at each epoch. ``Default'' is a standard training method, and ``MORPH'' is our proposed one; (c) contrasts the convergence of their test error.} 
\label{fig:in_depth_analysis}
\vspace*{-0.4cm}
\end{figure*}

Although this family of methods has achieved better generalization by training with selected small-loss training samples, they commonly have the following \emph{two} problems:
\begin{enumerate}[label={\arabic*.}, leftmargin=9pt] 
\item \textbf{Discarding Hard Sample}: True-labeled samples with large losses are simply discarded by the small-loss trick, though they have a great impact on generalization\,\cite{chang2017active}. This issue can be exacerbated by real-world and asymmetric label noises, where the loss distributions of true- and false-labeled samples are overlapped closely\,\cite{song2020learning}.
\vspace*{0.05cm}
\item \textbf{Inefficient Learning}: The small-loss trick suffers from confirmation bias\,\cite{tarvainen2017mean}, which is a hazard of favoring the samples selected at the beginning of training. Hence, recent approaches often leverage multiple DNNs to cooperate with one another\,\cite{han2018co, yu2019does} or run multiple training rounds to iteratively refine their selected set of clean samples\,\cite{shen2019learning, song2019selfie}, thus adding heavy computational overhead.
\end{enumerate}

In this regard, we have thoroughly investigated the memorization of a DNN on {real-world} noisy training samples and, consequently, observed the existence of \emph{two} learning periods in Figure \ref{fig:in_depth_analysis}(a): \textbf{(i)} the {\enquote{noise-robust}} period where the memorization of false-labeled samples is insignificant because the DNN prefers memorizing easy samples at an early stage and \textbf{(ii)} the {\enquote{noise-prone}} period where the memorization of false-labeled samples rapidly increases because the DNN eventually begins memorizing all the noisy samples at a late stage of training.

These findings motivate us to come up with an approach that leverages the transitional memorization nature in a single training round. In this paper, we propose \textbf{MORPH}, which is a \emph{self-transitional} learning approach that automatically transitions its learning phase when a DNN enters the \enquote{noise-prone} period after the \enquote{noise-robust} period\,(i.e., the dashed line in Figure \ref{fig:in_depth_analysis}(b)). Thus, corresponding to these two periods, our key idea divides the training process into two learning phases, namely \emph{seeding} and \emph{evolution}:
\vspace*{-0.3cm}
\begin{enumerate}[label={\arabic*.}, leftmargin=9pt] 
\item \phaseI{}: 
Owing to the negligible memorization of false-labeled samples, the model update is initiated using \emph{all} the training samples in the noise-robust period. As the samples memorized at this time are mostly easy samples with true labels, they are accumulated as a \emph{clean} seed to derive a \emph{maximal safe set} in the next phase. Note that \algname{} automatically estimates the best phase transition point without \emph{any} supervision.
\vspace*{0.05cm}
\item \phaseII{}: 
Without memorizing false-labeled samples, the DNN evolves by being updated \emph{only} for the maximal safe set in the noise-prone period. Then, the updated DNN recognizes more true-labeled samples previously hard to distinguish and filters out false-labeled samples incorrectly included. This alternating process repeats \emph{per iteration} so that the maximal safe set is expanded and refined in the remaining noise-prone period. 
\end{enumerate}

\footnotetext{We used the subset in which correct labels are identified.}

Through self-transitional learning, \algname{} avoids the confirmation bias by exploiting the noise-robust period with all the training samples, thus eliminating the need for additional DNNs and training rounds.
In addition, it incrementally expands the clean seed towards the maximal safe set, which can cover even hard true-labeled samples, not just throwing them away. The alternating process in the second phase minimizes the risk of misclassifying false-labeled samples as true-labeled ones or vice versa. 
Hence, as shown in Figure \ref{fig:in_depth_analysis}(b), \algname{} prevents the memorization of false-labeled samples by training with the maximal safe set during the noise-prone period, while increasing the memorization of true-labeled samples. As a result, as shown in Figure \ref{fig:in_depth_analysis}(c), the generalization performance of a DNN improves remarkably even in real-world noise.  
Our main contributions are summarized as follows:
\begin{itemize}[leftmargin=9pt]
\item \textbf{No Supervision for Transition}: \algname{} performs \emph{self-transitional} learning without \emph{any} supervision such as a true noise rate and a clean validation set, which are usually hard to acquire in real-world scenarios.
\item \textbf{Noise Robustness}:
Compared with state-of-the-art methods, \algname{} identifies true-labeled samples from noisy data with much higher recall and precision. Thus, \algname{} improved the test\,(or validation) error by up to $27.0pp$\footnote{A $pp$ is the abbreviation of a percentage point.} for  three datasets with two synthetic noises and by up to $8.90pp$ and $3.85pp$ for WebVision 1.0 and FOOD-101N with real-world noise.
\vspace*{0.05cm}
\item \textbf{Learning Efficiency}: Differently from other methods, \algname{}  requires neither additional DNNs nor training rounds. Thus, it was significantly faster than others by up to $3.08$ times. 
\end{itemize}
\vspace*{0.1cm}
\section{Related Work}
\label{sec:related_work}
\vspace*{0.05cm}

Numerous approaches have been proposed to address the challenge of learning from noisy labels. For a more thorough study on this topic, we refer the reader to surveys\,\cite{frenay2013classification, song2020learning}.

\smallskip \noindent\textbf{Loss Correction}:
A typical method is using \enquote{loss correction,} which estimates the label transition matrix and corrects the loss of the samples in a mini-batch. 
{Bootstrap}\,\cite{reed2014training} updates the DNN based on its own reconstruction-based objective with the notion of perceptual consistency. {F-correction}\,\cite{patrini2017making} reweights the forward or backward loss of the training samples based on the label transition matrix estimated using a pre-trained DNN. 
{D2L}\,\cite{ma2018dimensionality} employs a simple measure called local intrinsic dimensionality and then uses it to modify the forward loss in order to reduce the effects of false-labeled samples in learning.
\citet{ren2018learning} included a small amount of clean validation data into the training data and reweighted the backward loss of the mini-batch samples such that the updated gradient minimized the loss of those validation data.
However, this family of methods accumulates severe noise from the \emph{false correction}, especially when the number of classes or the number of false-labeled samples is large\,\cite{han2018co}. 

\smallskip \noindent\textbf{Sample Selection}:
To be free from the false correction, many recent researches have adopted the method of \enquote{sample selection,} which trains the DNN on selected samples. This method attempts to select the true-labeled samples from the noisy training data for updating the DNN.
{Decouple}\,\cite{malach2017decoupling} maintains two DNNs simultaneously and updates the models by only using the samples with different label predictions from these two DNNs.
\citet{wang2018iterative} proposed an iterative learning framework that learns deep discriminative features from well-classified noisy samples based on the local outlier factor algorithm.
{MentorNet}\,\cite{jiang2017mentornet} introduces a collaborative learning paradigm in which a pre-trained mentor DNN guides the training of a student DNN. Based on the small-loss criteria, the mentor provides the student with the samples whose labels are probably correct. 
{Co-teaching}\,\cite{han2018co} and {Co-teaching+}\,\cite{yu2019does} also maintain two DNNs, but each DNN selects a certain number of small-loss samples and feeds them to its peer DNN for further training. Compared with {Co-teaching}, {Co-teaching+} further employs the disagreement strategy of {Decouple}. 
{INCV}\,\cite{chen2019understanding} randomly divides the noisy training data and then employs cross-validation to classify true-labeled samples while removing large-loss samples at each training round.
{ITLM}\,\cite{shen2019learning} iteratively minimizes the trimmed loss by alternating between selecting a fraction of small-loss samples at the current moment and retraining the DNN using them. {SELFIE}\,\cite{song2019selfie} trains a DNN on selectively refurbished samples together with small-loss samples. 
However, this family of methods suffers from the two challenges: \textbf{(i}) hard training samples with true labels are simply discarded; and \textbf{(ii)} the additional overhead for multiple DNNs or training rounds hinders them from being scalable to larger problems. \looseness=-1

\smallskip \noindent\textbf{Other Directions}:
Beyond the supervised learning scope, Meta-Weight-Net\,\cite{shu2019meta} employs a meta-learning approach to reweight the loss of the training samples. Some other recent studies have attempted to combine unsupervised and semi-supervised learning. DM-DYR-SH\,\cite{arazo2019unsupervised} corrects the loss of training samples by modeling label noise in an unsupervised manner. {DivideMix}\,\cite{li2020dividemix} treats small-loss training samples as labeled ones to adopt a semi-supervised learning technique called {MixMatch}\,\cite{berthelot2019mixmatch}. 

\smallskip \noindent\textbf{Difference from Existing Work}:
Among these directions, our approach belongs to \enquote{sample selection} and is easily combined with meta-learning, unsupervised learning, and semi-supervised learning approaches because they are all \emph{orthogonal} to ours. The main technical novelty of \algname{} is to train the DNN with the maximal safe set, which is safely expanded from the clean seed through the notion of self-transitional learning. 
Furthermore, compared with prior studies on the memorization of DNNs\,\cite{arpit2017closer, liu2020early, li2020gradient}, this is the \emph{first} work to solve the problem of finding the best transition point.
\section{Preliminary}
\label{sec:preliminaries}

A $k$-class classification problem requires training data $\mathcal{D}=\{(x_i, y_i^{*})\}_{i=1}^{N}$, where $x_i$ is a sample and $y_i^{*}\in$ $\{1,2,\dots,k\}$ is its \emph{true} label. Following the label noise scenario, let's consider the noisy training data $\tilde{\mathcal{D}}=\{(x_i, \tilde{y}_i)\}_{i=1}^{N}$, where $\tilde{y}_i\in\{1,2,\dots,k\}$ is a \emph{noisy} label which may not be true. Moreover, in conventional training, a mini-batch $\mathcal{B}_{t}=\{(x_i, \tilde{y}_i)\}_{i=1}^{b}$ comprises $b$ samples drawn randomly from the noisy training data $\tilde{\mathcal{D}}$ at time $t$. Given a DNN parameterized by $\Theta$, the DNN parameter $\Theta$ is updated according to the decent direction of the expected loss on the mini-batch via stochastic gradient descent,
\vspace{-0.1cm}
\begin{equation}
\label{eq:corrupted_update}
\Theta_{t+1} = \Theta_{t} - \eta\nabla\Big(\frac{1}{|\mathcal{B}_{t}|}\!\sum_{(x,\tilde{y}) \in \mathcal{B}_{t}} \!\!\!\!f_{(x, \tilde{y})}(\Theta_t)\Big),
\end{equation}
where $\eta$ is the given learning rate and $f_{(x, \tilde{y})}(\Theta_t)$ is the loss of the sample $(x, \tilde{y})$ for the DNN parameterized by $\Theta_t$.

As for the notion of DNN memorization, a sample $x$ is defined to be \emph{memorized} by a DNN if the majority of its recent predictions at time $t$ coincide with the given label, as in Definition \ref{def:fully_learned_sample}. \looseness=-1

\begin{definition}\textbf{(Memorized Sample)}
\label{def:fully_learned_sample}
Given a DNN $\Phi$ parameterized by $\Theta$, let $\hat{y}_{t}=\Phi(x;\Theta_{t})$ be the predicted label of a sample $x$ at time $t$ and $\mathcal{H}_{x}^{t}(q)=\{\hat{y}_{t_1},\hat{y}_{t_2},\dots,\hat{y}_{t_q}\}$ be the history of the sample $x$ that stores the predicted labels of the recent $q$ epochs.
Next, based on $\mathcal{H}_{x}^{t}(q)$, the probability of a label $y\in\{1,2,\dots k\}$ estimated as the label of the sample $x$ is formulated by
\begin{equation}
\label{eq:label_prob}
\begin{gathered}
p(y|x,t) = \frac{1}{{|\mathcal{H}_{x}^{t}(q)|}}{\sum_{\hat{y} \in \mathcal{H}_{x}^{t}(q)}[\hat{y}=y]},\\
\text{where}~~ [S] = \begin{cases} 1, & \text{if $S$ is true}. \\ 0, & \text{otherwise.} \end{cases}
\end{gathered}
\end{equation}
Subsequently, the sample $x$ with its noisy label $\tilde{y}$ is defined to be \emph{memorized} by the DNN with the parameter $\Theta_{t}$ at time $t$ if ${\rm argmax}_{y}p(y|x,t) = \tilde{y}$ holds. \qed
\end{definition}

\section{\algname{}: Self-Transitional Learning}
\label{sec:methodology}

In this section, we propose \algname{} which comprises the following two phases: \textbf{Phase I} for preparing the clean seed during the noise-robust period and \textbf{Phase II} for carefully evolving it towards the maximal safe set during the noise-prone period. 

\subsection{Phase I: \phaseI{}} 
\label{sec:phase1}
Phase I initiates to update the DNN using \emph{all} the training samples in a conventional way of Eq.\,\eqref{eq:corrupted_update} during the noise-robust period, where the memorization of false-labeled samples is suppressed. 
Concurrently, because most of the samples memorized until the transition point are true-labeled, \algname{} collects them to form a clean \emph{seed}, which is used as an initial maximal safe set in Phase II. The major challenge here is to estimate the best transition point. 

\subsubsection{{Best Transition Point}}
The DNN predominantly learns true-labeled samples until the noise-prone period begins. 
That is, at the best phase transition point, the DNN \textbf{(i)} not only acquires \emph{sufficient} information from the true-labeled samples, \textbf{(ii)} but also accumulates \emph{little} noise from the false-labeled ones. In that sense, we propose two memorization metrics, namely, \emph{memorization recall\,(MR)} and \emph{memorization precision\,(MP)} in Definition \ref{def:memorization_metrics}, which are indicators of evaluating the two properties, respectively. 

\begin{definition}\textbf{(Memorization Metrics)}
\label{def:memorization_metrics}
Let $\mathcal{M}_t\subseteq\tilde{\mathcal{D}}$ be a set of memorized samples at time $t$ according to Definition \ref{def:fully_learned_sample}.  Then, \emph{memorization recall} and \emph{precision} at time $t$ are formulated by
\begin{equation}
\begin{gathered}
{MR(t)} = \frac{|\{(x,\tilde{y})\in\mathcal{M}_t:\tilde{y}=y^{*}\}|}{|\{(x,\tilde{y})\in\tilde{\mathcal{D}}:\tilde{y}=y^{*}\}|},\\
{MP(t)} = \frac{|\{(x,\tilde{y})\in\mathcal{M}_t:\tilde{y}=y^{*}\}|}{|\mathcal{M}_t|}
\end{gathered}
\label{eq:memorization_metrics}
\end{equation} 
$\text{where}~~\mathcal{M}_t=\{(x,\tilde{y}) \in \tilde{\mathcal{D}} : \text{argmax}_{y}p(y|x,t)=\tilde{y}\}.$ \qed
\end{definition}

By Definition \ref{def:memorization_metrics}, the best transition point $t_{tr}$ is naturally the moment when the recall and precision exhibit the highest value at the same time. This decision is also supported by the empirical understanding of memorization in deep learning that a better generalization of a DNN is achieved when pure memorization\,(i.e., high MP) and its enough amount\,(i.e., high MR) are satisfied simultaneously\,\cite{zhang2019identity, li2020gradient}. However, it is not straightforward to find the best transition point without the ground-truth labels $y^{*}$. 

\begin{figure}[t!]
\begin{center}
\includegraphics[width=8.55cm]{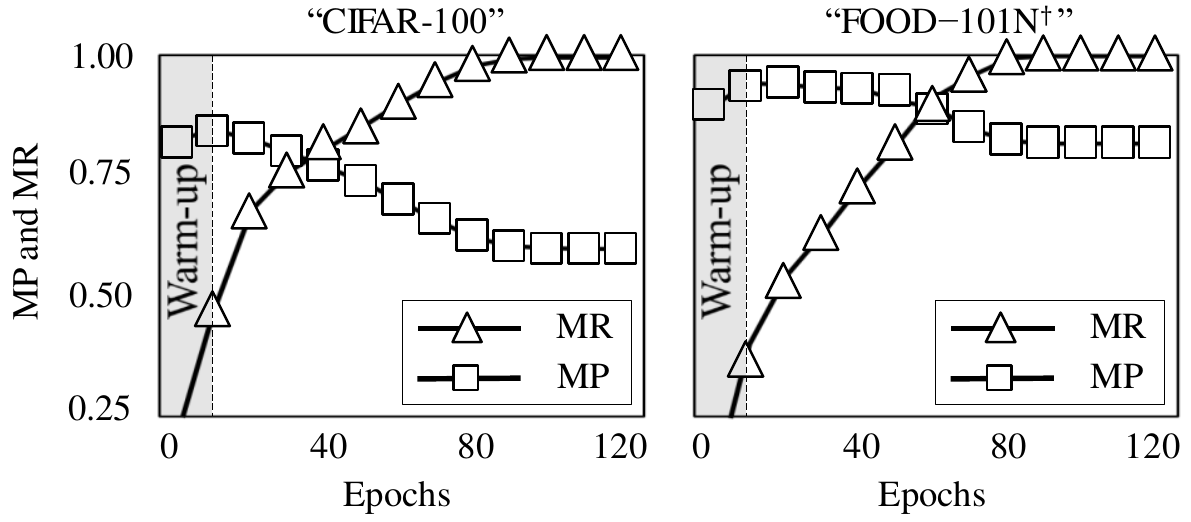}
\end{center}
\vspace*{-0.13cm}
\hspace*{0.52cm} {\small (a) Asymmetric Noise.} \hspace*{1.5cm} {\small (b) Real Noise.}
\vspace*{-0.25cm}
\caption{MR and MP when training WideResNet-16-8 on CIFAR-100 with the asymmetric noise of $40\%$ and $\text{FOOD-101N}^{\dag}$ with the real-world noise of $18.4\%$, where $\dag$ indicates the subset in which correct labels are identified.}
\label{fig:monotonicity}
\vspace*{-0.3cm}
\end{figure}

Figure \ref{fig:monotonicity} shows the change in MR and MP on the two \emph{realistic} label noises over the training period. These two metrics were observed to follow the \emph{monotonicity} after a few warm-up epochs for the following reasons\,(see Section \ref{sec:theory} for theoretical analysis):
\begin{itemize}[leftmargin=9pt] 
\item 
\emph{MR} monotonically increases because a DNN eventually memorizes all the true-labeled samples as the training progresses.
\item 
\emph{MP} monotonically decreases because a DNN tends to memorize true-labeled samples first and then gradually memorizes all the false-labeled samples.
\end{itemize} 

Under the monotonicity, the best transition point is the one \emph{cross-point} of the two metrics, i.e., $MP(t)=MR(t)$, because it is the \emph{best trade-off} between them. Then, by finding the answer of $MP(t)=MR(t)$ in Eq.\ \eqref{eq:memorization_metrics}, the best transition point $t_{tr}$ satisfies
\begin{equation}
\begin{gathered}
|\mathcal{M}_{t_{tr}}| = |\{(x,\tilde{y})\in\tilde{\mathcal{D}}:\tilde{y}=y^{*}\}|= (1-\tau)|\tilde{\mathcal{D}}|\\
\therefore |\mathcal{M}_{t_{tr}}| = (1-\tau)|\tilde{\mathcal{D}}|,
\end{gathered}
\label{eq:cross_condition}
\end{equation}
where $\tau$ is the true noise rate. 
Because $\tau$ is typically unknown, it is \emph{automatically} estimated by \algname{} to check the condition in Eq.\,\eqref{eq:cross_condition} for the phase transition.

Regarding the noise rate estimation, \algname{} fits a two-component Gaussian Mixture Model\,(GMM) to model the loss distribution of true- and false-labeled samples because the distribution is \emph{bi-modal} \cite{arazo2019unsupervised, pleiss2020detecting}. At each epoch, \algname{} accumulates the loss of all the training samples and fits the GMM to the accumulated loss by using the Expectation-Maximization\,(EM) algorithm. The probability of a sample $x$ being false-labeled is obtained through its posterior probability. Accordingly, the noise rate $\tau$ is estimated by 
\begin{equation}
\hat{\tau} =\mathbb{E}_{(x,\tilde{y})\in\tilde{\mathcal{D}}}[p\big(g|f_{(x, \tilde{y})}(\Theta_t)\big)],
\label{eq:estimated_noise_rate}
\end{equation}
where $g$ is the Gaussian component with a larger mean\,(i.e., larger loss). A thorough analysis of estimating the noise rate is provided in Appendix \ref{sec:generalization}.  

Therefore, \algname{} transitions the learning phase at time $t_{tr}$ when the number of memorized samples is greater than or equal to the estimated number of true-labeled ones\,(i.e., $|\mathcal{M}_{t_{tr}}| \geq (1-\hat{\tau})|\tilde{\mathcal{D}}|$). Please note that \algname{} requires \emph{neither} a true noise rate \emph{nor} a clean validation set. 

\subsubsection{{Theoretical Analysis}}
\label{sec:theory}

We formally prove that MR and MP change monotonically over the training time $t$. Let us make an assumption that a set of memorized samples $\mathcal{M}_{t}$ grows as a loss value decreases, which generally holds because the loss value implies how well a certain DNN predicts the label of training samples.
\begin{assumption}
\label{assum:set_size}
Let $f_{\mathcal{\tilde{D}}}(\Theta_t)$ be the expected loss of the DNN parameterized by $\Theta_t$ for all $(x,\tilde{y}) \in \mathcal{\tilde{D}}$. Then, it is assumed that
\begin{equation}
f_{\mathcal{\tilde{D}}}(\Theta_t) \geq f_{\mathcal{\tilde{D}}}(\Theta_{t+1}) \rightarrow |\mathcal{M}_t| \leq |\mathcal{M}_{t+1}|~{\rm and}~ \mathcal{M}_t \subseteq \mathcal{M}_{t+1}. \!\!\! \qed
\label{eq:inverse}
\end{equation}
\end{assumption}

Subsequently, the monotonicity of MR and MP is proven by Theorems \ref{theorem:mr} and \ref{theorem:mp} under Assumption \ref{assum:set_size}.

\begin{theorem}
Let the DNN be trained with gradient descent with a learning rate $\eta$. Then, $MR(t)$ is a monotonically increasing function that converges to $1$ over the training time $t$.
\vspace*{-0.2cm}
\begin{proof}\phantom{\qedhere} 
\!\!\!\!\!\!\!\!\!\!\!\!Suppose that the loss function $f: \mathbb{R}^{d} \rightarrow \mathbb{R}$ be a L-Lipschitz convex function. 
Then, the loss value monotonically decreases with gradient descent\,\cite{shamir2013stochastic}, 
\begin{equation}
f_{\mathcal{\tilde{D}}}(\Theta_{t+1}) ~~\leq~~ f_{\mathcal{\tilde{D}}}(\Theta_{t}) - \frac{\eta}{2}||\nabla f_{\mathcal{\tilde{D}}}(\Theta_{t})||_2^2 ~~\leq~~ f_{\mathcal{\tilde{D}}}(\Theta_{t}).
\end{equation}
Hence, by Assumption \ref{assum:set_size}, $|\mathcal{M}_t|$ increases monotonically from the update. That is, $\mathcal{M}_t$ gradually includes more true-labeled samples and eventually contains all of them. Thus, Eq.~\eqref{eq:mr} holds.
\begin{equation}
MR(t) \leq MR(t+1)~ {\rm and}~ {\rm lim}_{t \to \infty}MR(t)=1. \qed
\label{eq:mr}
\end{equation}
\end{proof}
\label{theorem:mr}
\end{theorem}

\vspace*{-0.35cm}
\begin{theorem}
Let the DNN be trained with gradient descent with a learning rate $\eta$. Then, $MP(t)$ is a monotonically decreasing function that converges to $(1-\tau)$ over the training time $t$, where $\tau \in [0, 1]$ is the ratio of false-labeled samples in the training data.
\label{theorem:mp}
\vspace*{-0.2cm}
\begin{proof}\phantom{\qedhere} 
\!\!\!\!\!\!\!\!\!\!\!\! Let $\mathcal{M}_t = \mathcal{C}_t \cup \mathcal{R}_t$, where $\mathcal{C}_t$ and $\mathcal{R}_t$ are the sets of true- and false-labeled samples, respectively, memorized at time $t$. Then, ${|\mathcal{R}_{t+1}|}/{|\mathcal{R}_{t}|} \geq {|\mathcal{C}_{t+1}|}/{|\mathcal{C}_{t}|}$ because false-labeled samples are memorized rapidly as the training process progresses contrary to true-labeled samples after the DNN stabilizes by a few epochs\,\cite{arpit2017closer}.  
Hence, $MP(t) \geq MP(t+1)$ through the following derivation,
\begin{equation}
\begin{split}
& {|\mathcal{R}_{t+1}|}~/~{|\mathcal{R}_{t}|} \geq {|\mathcal{C}_{t+1}|}~/~{|\mathcal{C}_{t}|}   \\
&\Leftrightarrow   {|\mathcal{R}_{t+1}|}{|\mathcal{C}_{t}|} + {|\mathcal{C}_{t+1}|}{|\mathcal{C}_{t}|} \geq {|\mathcal{C}_{t+1}|}{|\mathcal{R}_{t}|} + {|\mathcal{C}_{t+1}|}{|\mathcal{C}_{t}|} \\
&\Leftrightarrow   {|\mathcal{C}_{t}|}{(|\mathcal{C}_{t+1}| + |\mathcal{R}_{t+1}|)} \geq {|\mathcal{C}_{t+1}|}{(|\mathcal{C}_{t}| + |\mathcal{R}_{t}|)}  {\color{white} spacing spacing} \!\!\\
&\Leftrightarrow   {|\mathcal{C}_{t}|}~/~|\mathcal{M}_{t}| \geq {|\mathcal{C}_{t+1}|}~/~|\mathcal{M}_{t+1}| \Leftrightarrow  MP(t) \geq MP(t+1). 
\end{split} 
\label{eq:derivation_mr}
\end{equation}
By gradient descent under Assumption \ref{assum:set_size}, $\mathcal{M}_t$ eventually contains all the false-labeled samples of size  $\tau|\mathcal{\tilde{D}}|$. Thus, Eq.~\eqref{eq:mp} holds.
\begin{equation}
MP(t) \geq MP(t+1)~ {\rm and}~ {\rm lim}_{t \to \infty}MP(t)=(1-\tau).  \qed
\label{eq:mp}
\end{equation}
\end{proof}
\end{theorem}

\vspace*{-0.25cm}
\subsection{Phase II: \phaseII{}} 
\label{sec:phase2}

Phase II robustly updates the DNN \emph{only} using the selected clean samples called the \emph{maximal safe set} in Definition \ref{def:maximal_safe_set}. By the definition of the transition point, the initial maximal safe set obtained in Phase I is quantitatively sufficient and qualitatively clean\,(i.e., high MR and MP); however, it can be further improved by \textbf{(i)} including more hard samples previously indistinguishable and \textbf{(ii)} filtering out false-labeled samples incorrectly included.
In this regard, we take advantage of the fact that \emph{undesired memorization} of false-labeled samples is easily forgotten by the robust update\,\cite{toneva2018empirical, han2020sigua}. Hence, \algname{} robustly updates the DNN with the current maximal safe set $\mathcal{S}_{t}$ as in Eq.~\eqref{eq:safe_update} and then derives a more refined set $\mathcal{S}_{t+1}$ as in Eq.~\eqref{eq:next_safe_set}. 

Thus, the DNN keeps even more hard true-labeled samples $\mathcal{C}_{new}$ owing to the robust update, while forgetting the false-labeled samples $\mathcal{R}_{new}$ which were incorrectly memorized earlier. In this way, \algname{} becomes better generalized to almost all true-labeled samples through the evolution of the safe set.

\vspace*{-0.05cm}
\begin{definition}\textbf{(Maximal Safe Set)}
\label{def:maximal_safe_set}
A \emph{maximal safe set} $\mathcal{S}_{t}$ is defined as the set of samples expected to be true-labeled at time $t (\geq t_{tr})$, where $t_{tr}$ is the transition point. It starts from $\mathcal{S}_{t_{tr}}=\mathcal{M}_{t_{tr}}$, and is managed as follows:

\begin{enumerate}[label={\arabic*.}, leftmargin=12pt] 
\item 
The refinement of the maximal safe set $\mathcal{S}_{t}$ accompanies the update of the DNN parameter $\Theta_{t}$.
To be free from memorizing false-labeled samples in Phase II, the DNN parameter $\Theta_{t+1}$ is robustly learned only using the samples in $\mathcal{S}_t$ out of the mini-batch $\mathcal{B}_t$ by
\begin{equation}
\label{eq:safe_update}
\Theta_{t+1} = \Theta_{t} - \alpha\nabla\big(\frac{1}{|\mathcal{B}_t \cap  \mathcal{S}_{t}|}\sum_{(x,\tilde{y}) \in (\mathcal{B}_t \cap  \mathcal{S}_{t})} \!\!\!\!\!\!\!\!\!f_{(x, \tilde{y})}(\Theta_t)\big). 
\end{equation} 
\item
Then, $\mathcal{S}_{t+1}$ reflects the changes by the DNN update at time $t+1$, i.e., newly memorized samples $\mathcal{C}_{new}$ and newly forgotten samples $\mathcal{R}_{new}$, as formulated by 
\begin{equation}
\label{eq:next_safe_set}
\begin{gathered}
\mathcal{S}_{t+1} = \mathcal{S}_{t} + \mathcal{C}_{new} - \mathcal{R}_{new}, ~~\text{where}\\
\!\!\!\!\!\!\!\!\!\!\!\!\mathcal{C}_{new} = \{(x,\tilde{y})\in(\mathcal{B}_{t}\cap\mathcal{S}_{t}^{c}):{\rm argmax}_{y}p(y|x,t+1) = \tilde{y}\},\!\!\!\\
\,\,\,\,\!\!\!\!\!\mathcal{R}_{new} = \{(x,\tilde{y})\in{(\mathcal{B}_{t}\cap\mathcal{S}_{t})}:{\rm argmax}_{y}p(y|x,t+1) \neq \tilde{y}\}. \!\!\qed 
\end{gathered}
\end{equation}
\end{enumerate}
\end{definition}

Furthermore, to avoid the potential risk of overfitting to a small \emph{initial} seed, which is typically observed with a very high noise rate, \algname{} adds \emph{consistency regularization} $\mathcal{J}(\Theta_t)$\,\cite{laine2017temporal, zhang2019consistency} to the supervised loss in Eq.\,\eqref{eq:safe_update}.
Without relying on possibly unreliable labels, this regularization effectively helps learn the \emph{dark knowledge}\,\cite{hinton2015distilling} from \emph{all} the training samples by penalizing the prediction difference between the original sample $x$ and its augmented sample $\hat{x}$.\,(For the experiments, $\hat{x}$'s were generated by random crops and horizontal flips.) 
Hence, the update rule is finally defined by
\begin{equation}
\begin{gathered}
\label{eq:modified_safe_update}
\Theta_{t+1} = \Theta_{t}-\alpha\nabla\big(\frac{1}{|\mathcal{B}_t \cap  \mathcal{S}_{t}|}\!\!\!\!\!\!\!\!\sum_{\,\,\,\,\,\,\,\,\,(x, \tilde{y}) \in (\mathcal{B}_t \cap  \mathcal{S}_{t})}\!\!\!\!\!\!\!\!\!\! f_{(x, \tilde{y})}(\Theta_t) + w(t)\mathcal{J}(\Theta_t)\big), \\\text{where}~
\mathcal{J}(\Theta_{t})= \frac{1}{\mathcal{B}_{t}}\sum_{(x,\tilde{y}) \in \mathcal{B}_t}||z(x;\Theta_{t}) - z({\hat{x}};\Theta_{t})||_{2}^{2}, 
\end{gathered}
\end{equation}
where $z(x;\Theta_{t})$ is the softmax output of a sample $x$ and $w(t)$ is a Gaussian ramp-up function to gradually increase the weight to the maximum value $w_{max}$. 
According to our ablation study in Section \ref{sec:consistency}, the regularization further enhanced the robustness of MORPH when dealing with very high noise rates.

\setlength{\textfloatsep}{10pt}
\begin{algorithm}[t!]
\caption{MORPH}
\label{alg:proposed_algorithm}
\begin{algorithmic}[1]
\REQUIRE {$\tilde{\mathcal{D}}$: data, $epochs$: total number of epochs, $q$: history length, $w_{max}$: maximum weight for $\mathcal{J}$}
\ENSURE {$\Theta_t$: DNN parameters, $\mathcal{S}_{t}$: final safe set}
\INDSTATE[0] {$t \leftarrow 1$; ~$\mathcal{S}_t \leftarrow \emptyset$; ~$\Theta_{t} \leftarrow \text{Initialize DNN parameters};$}
\INDSTATE[0] {\color{blue}\COMMENT{{\bf I.~\phaseI{} during Noise-Robust Period}}}
\INDSTATE[0] {\bf for} $i=1$ {\bf to} $epochs$ {\bf do} 
\INDSTATE[1] {\bf for} $j=1$ {\bf to} $|{\tilde{\mathcal{D}}}|/|\mathcal{B}_t|$ {\bf do}
\INDSTATE[2] {Draw a mini-batch $\mathcal{B}_t$ from $\tilde{\mathcal{D}}$;}
\INDSTATE[2] \COMMENT{Standard update by Eq.\,(\ref{eq:corrupted_update})}
\INDSTATE[2] $\Theta_{t+1} = \Theta_{t} - \alpha\nabla \big(\frac{1}{|\mathcal{B}_t|} \sum_{(x,\tilde{y}) \in \mathcal{B}_t} f_{(x,\tilde{y})}(\Theta_t)\big)$;
\INDSTATE[2] $t \leftarrow t+1;$
\INDSTATE[1] \COMMENT{Noise rate estimation by Eq.\,(\ref{eq:estimated_noise_rate})}
\INDSTATE[1] $\hat{\tau} \leftarrow \mathbb{E}_{(x,\tilde{y})\in\tilde{\mathcal{D}}}[p\big(g|f_{(x, \tilde{y})}(\Theta_t)\big)];$
\INDSTATE[1] \COMMENT{Checking the phase transition condition}
\INDSTATE[1] {\bf if} $|\mathcal{M}_{t}| \geq (1-\hat{\tau})|\tilde{\mathcal{D}}|$ {\bf then}
\INDSTATE[2] \COMMENT{Assigning the initial maximal safe set}
\INDSTATE[2] $t_{tr} \leftarrow t$; ~ $\mathcal{S}_{t_{tr}} \leftarrow \mathcal{M}_{t_{tr}}$; break;
\INDSTATE[0] {\color{blue}\COMMENT{{\bf II.~\phaseII{} during Noise-Prone Period}}}
\INDSTATE[0] {\bf for} $i=t_{tr}|\mathcal{B}_t|/|{\tilde{\mathcal{D}}}|+1$ {\bf to} $epochs$ {\bf do} 
\INDSTATE[1] {\bf for} $j=1$ {\bf to} $|{\tilde{\mathcal{D}}}|/|\mathcal{B}_t|$ {\bf do}
\INDSTATE[2] Draw a mini-batch $\mathcal{B}_t$ from $\tilde{\mathcal{D}}$; 
\INDSTATE[2] \COMMENT{Robust update by Eq.\,(\ref{eq:modified_safe_update})}
\INDSTATE[2] $\Theta_{t+1}=\Theta_{t}$ $- \alpha\nabla\big(\frac{1}{|\mathcal{B}_t \cap  \mathcal{S}_{t}|}$$\sum_{(x, \tilde{y}) \in (\mathcal{B}_t \cap  \mathcal{S}_{t})}$$f_{(x, \tilde{y})}(\Theta_t)$ \\\text{\,\,\,\,\,\,\,\,\,\,\,\,\,\,\,\,\,\,\,\,\,\,\,\,\,\,\,\,\,\,\,\,\,\,\,\,\,\,\,\,\,\,\,\,\,\,\,\,\,\,\,} $+ w(t)\mathcal{J}(\Theta_t)\big)$;
\INDSTATE[2] \COMMENT{Updating the maximal safe set by Eq.\,\eqref{eq:next_safe_set}}
\INDSTATE[2] $\mathcal{S}_{t+1} \leftarrow \mathcal{S}_t + \mathcal{C}_{new} - \mathcal{R}_{new};$
\INDSTATE[2] $t \leftarrow t+1;$
\INDSTATE[0] {\bf return} $\Theta_{t}$, $\mathcal{S}_{t}$;
\end{algorithmic}
\end{algorithm} 

\vspace*{-0.05cm}
\subsection{Algorithm Pseudocode}
\label{sec:algorithm}

Algorithm \ref{alg:proposed_algorithm} describes the overall procedure of \algname{}, which is self-explanatory.
First, the DNN is trained on the noisy training data $\tilde{\mathcal{D}}$ in the \emph{default} manner\,(Lines\,$5$--$7$). During the first phase, the noise rate is estimated and subsequently used to find the moment when the {noise-prone} period begins. Here, if the transition condition holds, Phase I transitions to Phase II after assigning the clean seed\,(Lines\,$9$--$14$). 
Subsequently, during the second phase, the mini-batch samples in the current maximal safe set are selected to update the DNN parameter with the consistency regularization\,(Lines\,$18$--$20$). The rest mini-batch samples are excluded to pursue the robust learning. Finally, the maximal safe set is refined by reflecting the change in DNN memorization resulting from the update\,(Lines $21$--$22$). This alternating process is repeated until the end of the remaining learning period.

{
\begin{figure*}[!t]
\begin{center}
{\includegraphics[width=14cm]{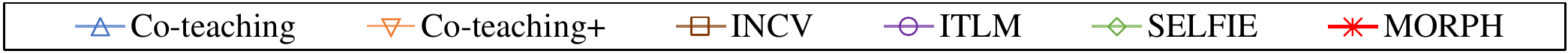}}
\end{center}
\vspace*{-0.1cm}
\begin{center}
{\includegraphics[height=36.5mm]{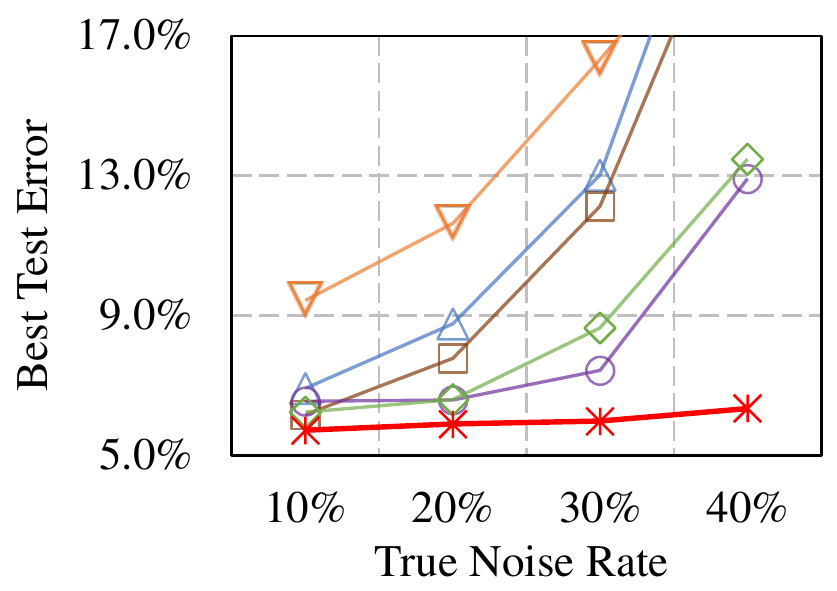}}
\hspace{0.5cm}
{\includegraphics[height=36.5mm]{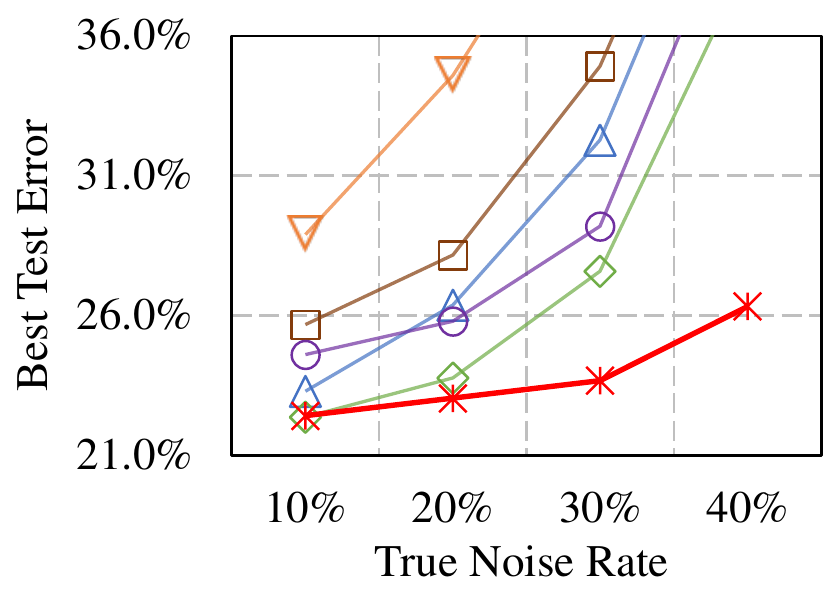}}
\hspace{0.5cm}
{\includegraphics[height=36.5mm]{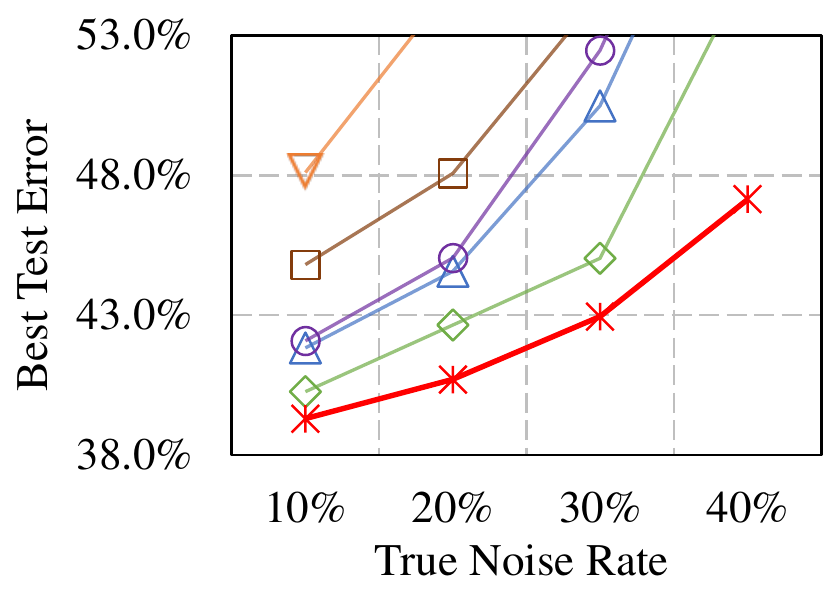}}
\end{center}
\vspace*{-0.1cm}
\hspace*{1.6cm} \small{(a) CIFAR-10.} \hspace*{4.1cm} \small{(b) CIFAR-100. \hspace*{3.75cm} \small{(c) Tiny-ImageNet.}}
\vspace*{-0.27cm}
\caption{Best test errors on three datasets using WideResNet with varying {asymmetric noise} rates.}
\vspace*{-0.5cm}
\label{fig:wrn_error_pair}
\end{figure*}
\begin{figure*}[!t]
\begin{center}
\includegraphics[height=36.5mm]{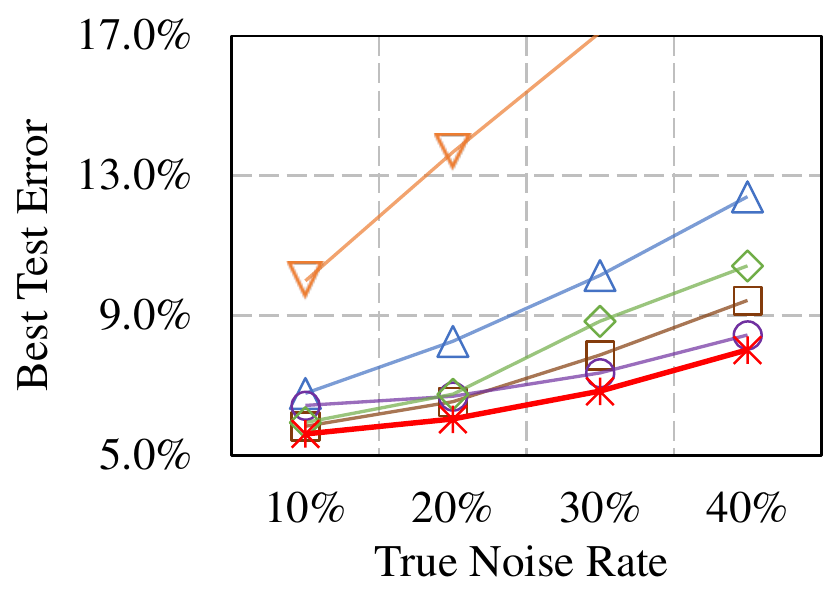}
\hspace{0.5cm}
\includegraphics[height=36.5mm]{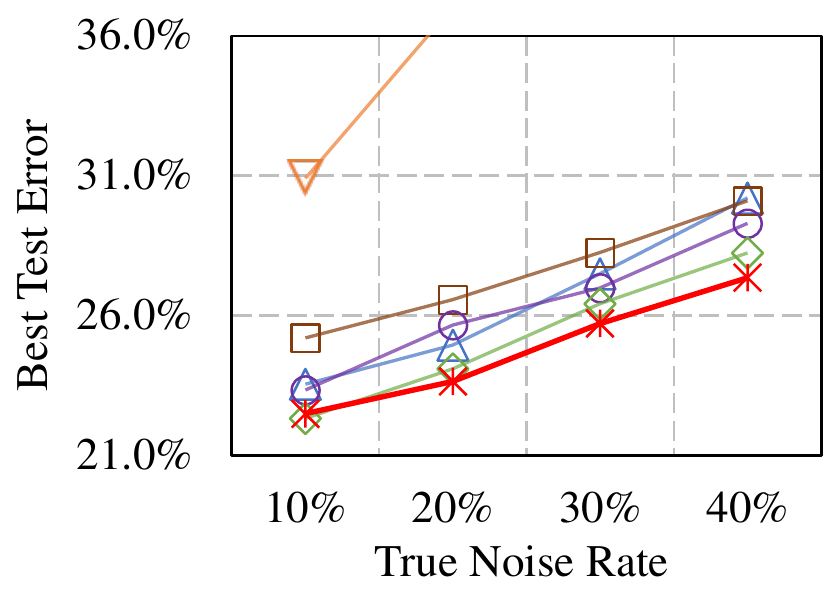}
\hspace{0.5cm}
\includegraphics[height=36.5mm]{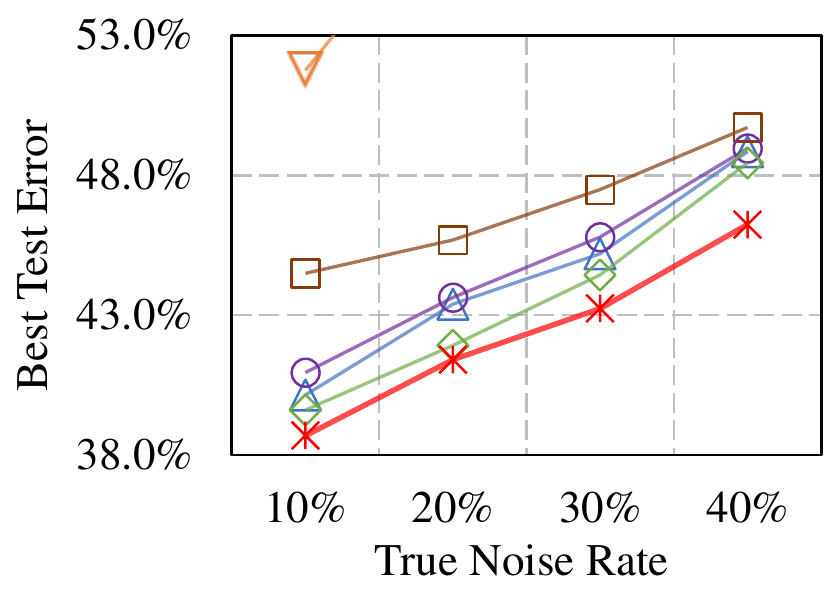}
\end{center}
\vspace*{-0.1cm}
\hspace*{1.6cm} \small{(a) CIFAR-10.} \hspace*{4.1cm} \small{(b) CIFAR-100. \hspace*{3.75cm} \small{(c) Tiny-ImageNet.}}
\vspace*{-0.27cm}
\caption{Best test errors on three datasets using WideResNet with varying {symmetric noise} rates.}
\label{fig:wrn_error_symmetric}
\vspace*{-0.3cm}
\end{figure*}
}

The main \emph{additional} costs of \algname{} are \textbf{(i)} the estimation of the noise rate\,(Line 10) and \textbf{(ii)} the additional inference step for the consistency regularization (Line 20). Because the noise rate is estimated using the EM algorithm once per epoch, its cost is negligible compared with that of the inference steps of a complex DNN. Thus, the additional inference in Phase II is the only part that increases the time complexity. Nevertheless, the additional cost is relatively \emph{cheap} considering that other sample selection methods\,\cite{han2018co, yu2019does,chen2019understanding, shen2019learning, song2019selfie} require either multiple additional DNNs or multiple training rounds.  



\vspace*{-0.1cm}
\section{Evaluation}
\label{sec:evaluation}
\vspace*{-0.1cm}

Our evaluation was conducted to support the followings:
\begin{itemize}[leftmargin=9pt]
\item \algname{} is \textbf{much robust} and \textbf{efficient} than other state-of-the-art sample selection methods\,(Section \ref{sec:robust_comparison} and \ref{sec:efficiency_comparison}).
\item \algname{} consistently identifies true-labeled samples from noisy data with \textbf{high recall} and \textbf{precision}\,(Section \ref{sec:robustness_comp_details}).
\item The transition point estimated by \algname{} is the \textbf{optimal point} for the best test error\,(Section \ref{sec:optimality}).
\item The consistency regularization is \textbf{much useful} at a very high noise rate\,(Section \ref{sec:consistency}).
\end{itemize}

\vspace*{-0.2cm}
\subsection{Experiment Setting}

\subsubsection{{Datasets}}
\label{sec:datasets}

To verify the superiority of \algname{}, we performed an image classification task on {five} benchmark datasets: CIFAR-10 and CIFAR-100, a subset of 80 million categorical images\,\cite{krizhevsky2014cifar}; Tiny-ImageNet\,\cite{krizhevsky2012imagenet}, a subset of ImageNet ILSVRC12\,\cite{deng2009imagenet}; WebVision 1.0\,\cite{li2017webvision}, real-world noisy data of large-scale web images crawled from Flickr and Google Images search; and FOOD-101N\,\cite{lee2018cleannet}, real-world noisy data of crawled food images annotated by their search keywords in the Food-101 Taxonomy. Random crops and horizontal flips were applied for data augmentation. Please see Appendix \ref{sec:data_decription} for the details of the benchmark datasets.

\subsubsection{{Noise Injection}}
\label{sec:noise_injection}
As all the labels in CIFAR and Tiny-ImageNet are clean, we artificially corrupted the labels in these datasets with two synthetic noises\,\cite{han2018co, yu2019does, song2019selfie}. 
For $k$ classes, we applied the label transition matrix $\textbf{T}$: \textbf{(i)} \emph{{symmetric noise}}: $\forall_{j \neq i} \textbf{T}_{ij}=\frac{\tau}{k-1}$, where a true label is flipped into other labels with equal probability; \textbf{(ii)} \emph{{asymmetric noise}}: $\exists_{j \neq i}\textbf{T}_{ij}=\tau \wedge \forall_{k \neq i, k \neq j}\textbf{T}_{ik}=0$, where a true label is flipped into a certain label.
Here, $\textbf{T}_{ij}$ is the probability of the true label $i$ being flipped to the corrupted label $j$, and $\tau$ is the noise rate. 
It is known that asymmetric noise is more realistic than symmetry noise because labelers may make mistakes only within very similar classes\,\cite{song2019selfie}.
To evaluate the robustness on varying noise rates from light noise to heavy noise, we tested four noise rates $\tau \in  \{0.1, 0.2, 0.3, 0.4\}$. In contrast, we did not inject any label noise into WebVision 1.0 and FOOD-101N because they contain real label noise whose rate is estimated at $20.0\%$ and $18.4\%$\,\cite{song2020learning}.

\begin{table*}[!t]
\noindent
\begin{center}
\parbox{.55\textwidth}{%
\parbox{9.3cm}{
\caption{Comparison with state-of-the-art methods trained on Webvision 1.0. The value outside\,(inside) the parentheses denotes the top-1\,(top-5) classification error\,(\%) on the WebVision validation set and the ImageNet ILSVRC12 validation set. The results for baseline methods are borrowed from \cite{chen2019understanding}.}
\label{table:webvision}
\vspace*{-0.3cm}
}
\begin{tabular}{L{3.0cm} |X{2.6cm} |X{2.6cm}}\toprule
{Method} & {\!\!WebVision Val.\!\!} & {\!\!\!ILSVRC12 Val.\!\!\!}  \\\midrule 
\emph{F-correction}{\, \cite{patrini2017making}}\!\!\!\! & 38.88 \,\,(17.32) & 42.64 \,\,(17.64) \\
\emph{Decouple}{\, \cite{malach2017decoupling}}\!\!\!\! & 37.46 \,\,(15.26) & 41.74 \,\,(17.74) \\
\emph{Co-teaching}{\, \cite{han2018co}}\!\!\!\! & 36.42 \,\,(14.80) & 38.52 \,\,(15.30) \\
\emph{MentorNet}{\, \cite{jiang2017mentornet}}\!\!\!\! & 37.00 \,\,(18.60) & 42.20 \,\,(20.08) \\
\emph{D2L}{\, \cite{ma2018dimensionality}}\!\!\!\! & 37.32 \,\,(16.00) & 42.20 \,\,(18.64) \\
\emph{INCV}{\, \cite{chen2019understanding} }\!\!\!\! & 34.76 \,\,(14.66) & 38.40 \,\,(15.02) \\
\textbf{\algname{}} & \textbf{29.98 \,\,(11.33)} & \textbf{33.09 \,\,(12.62)} \\\bottomrule
\end{tabular}
}%
\parbox{.6\textwidth}{%
\parbox{8.0cm}{
\caption{Comparison with state-of-the-art methods trained on FOOD-101N. The value denotes the top-1 classification error\,(\%) on the FOOD-101 validation set. The results for baseline methods are borrowed from \cite{lee2018cleannet, li2020product}. $\dag$ indicates that extra clean\,(or verification) labels were used for supervision.} 
\label{table:food101n}
\vspace*{-0.375cm}
}
\begin{tabular}{L{4.5cm} |X{2.7cm}}\toprule
{Method} & {\!\!\!FOOD-101 Val.\!\!\!}   \\\midrule 
\emph{Cross-Entropy}{\, \cite{lee2018cleannet}}\!\!\!\! & 18.56 \\
\emph{Weakly Supervised}{\, \cite{zhuang2017attend}}\!\!\!\! & 16.57 \\
$\emph{CleanNet}\,(w_{hard})^{\dag}${\, \cite{lee2018cleannet}}\!\!\!\! & 16.53 \\
$\emph{CleanNet}\,(w_{soft})^{\dag}${\, \cite{lee2018cleannet}}\!\!\!\! & 16.05 \\
$\emph{Guidance Learning}^{\dag}${\, \cite{li2020product}}\!\!\!\! & 15.80 \\
\textbf{\algname{}} & \textbf{14.71} \\\bottomrule
\end{tabular}
}
\end{center}
\vspace*{-0.2cm}
\end{table*}

\subsubsection{{Hyperparameter Configuration}}

\algname{} requires two additional hyperparameters: the history length $q$ and the maximum regularization weight $w_{max}$. 
We used the best history length $q\!=\!10$ and maximum weight $w_{max}\!=\!5.0$, which were obtained via a grid search in Appendix \ref{sec:hyper}. The $w$ value gradually increased from $0$ to $w_{max}$ using a Gaussian ramp-up function as the DNN with a large $w$ gets stuck in a degenerate solution at the beginning\,\cite{laine2017temporal}.
Similarly, the hyperparameters of the compared algorithms were configured to the best values, as detailed in Appendix \ref{sec:compared_algorithm}.

\vspace*{-0.05cm}
\subsubsection{{Reproducibility}}
For CIFARs and Tiny-ImageNet, we trained a WideResNet-16-8 from scratch using SGD with a momentum of $0.9$, a batch size of $128$, a dropout of $0.1$, and a weight decay of $0.0005$. The DNN was trained for $120$ epochs with an initial learning rate of $0.1$, which was decayed with cosine annealing. 
 
Regarding WebVision 1.0 and FOOD-101N, which contain the real-world noise, we followed exactly the same configuration in the previous work. For WebVision 1.0, we trained an Inception ResNet-V2\,\cite{szegedy2017inception} from scratch for the first $50$ classes of the Google image subset\,\cite{chen2019understanding}; for FOOD-101N, we fine-tuned a ResNet-50 with the ImageNet pretrained weights for the entire training set\,\cite{lee2018cleannet}. Please see Appendix \ref{sec:training_configuration} for the details of the configurations.

All the algorithms were implemented using TensorFlow $2.1.0$ and executed using $16$ NVIDIA Titan Volta GPUs. 
Note that the compared algorithms were  re-implemented by us for fair comparison.
In support of reliable evaluation, we repeated every task \emph{thrice} and reported the average test\,(or validation) error as well as the average training time. The source code and trained models are publicly available at \url{https://github.com/kaist-dmlab/MORPH}. 

\vspace*{-0.1cm}
\subsection{{Robustness Comparison}}
\label{sec:robust_comparison}

\subsubsection{{Results with Synthetic Noise}}
We compared \algname{} with the {five} state-of-the-art sample selection methods using the small-loss trick: \emph{Co-teaching}\,\cite{han2018co}, \emph{Co-teaching+}\,\cite{yu2019does}, \emph{INCV}\,\cite{chen2019understanding}, \emph{ITLM} \cite{shen2019learning}, and \emph{SELFIE}\,\cite{song2019selfie}.  In addition, the comparison with \emph{DM-DYR-SH}\,\cite{arazo2019unsupervised} and \emph{DivideMix}\,\cite{li2020dividemix}, which are in fact not directly comparable with \algname{}, is discussed in Appendix \ref{sec:com_recent}.
Figures \ref{fig:wrn_error_pair} and \ref{fig:wrn_error_symmetric} show the test errors of the six sample selection methods with varying \emph{asymmetric} and \emph{symmetric} noise rates. See Appendix \ref{sec:table_error_synthetic} for the tabular reports.

\begin{figure}
\vspace*{-0.1cm}
\begin{center}
\includegraphics[width=8.5cm]{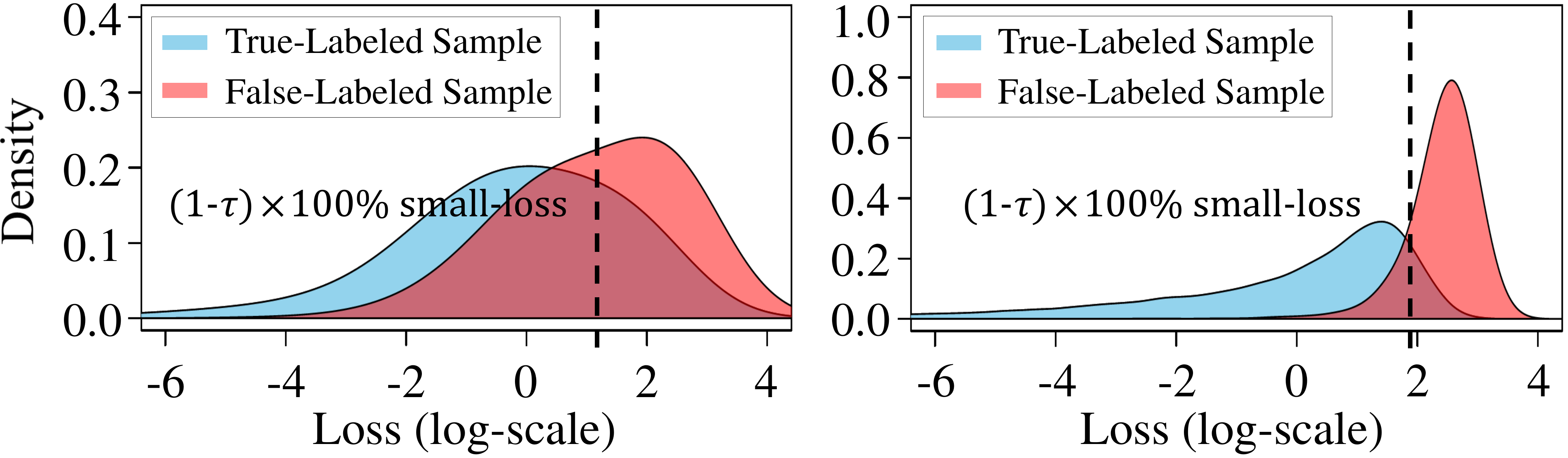}
\end{center}
\vspace*{-0.07cm}
\hspace*{0.65cm} {\small (a) Asymmetric Noise $40\%$.} \hspace*{1.0cm} {\small (b) Symmetric Noise $40\%$.}
\vspace*{-0.23cm}
\caption{Loss distribution of true- and false-labeled samples at the training accuracy of $50\%$ on noisy CIFAR-100.}
\label{fig:loss_distribution}
\vspace*{0.05cm}
\end{figure}

\smallskip \noindent\textbf{Asymmetric Noise:}
\algname{} generally achieved the lowest test errors with respect to a wide range of noise rates. The error reduction became larger as the noise rate increased, reaching $6.6pp$--$27.0pp$ at a heavy noise rate of 40\%. In contrast, the performance of the other methods worsened rapidly with an increase in the noise rate. As shown in Figure \ref{fig:loss_distribution}(a), because the loss distributions of true- and false-labeled samples are closely overlapped in asymmetric noise, the philosophy of selecting small-loss samples could not distinguish well true-labeled samples from false-labeled samples. Thus, the other methods simply discard a lot of hard training samples even if they are true-labeled. On the other hand, \algname{} gradually includes them by its alternating process in Phase II.

\smallskip \noindent\textbf{Symmetric Noise:}
\algname{} generally outperformed the other methods again, though the error reduction was relatively small, i.e., $0.42pp$--$26.3pp$ at a heavy noise rate of $40\%$. The small-loss trick was turned out to be appropriate for symmetric noise because the loss distributions are clearly separated, as illustrated in Figure \ref{fig:loss_distribution}(b).

\smallskip 
Putting them together, we contend that only \algname{} realizes \emph{noise type robustness}, evidenced by consistently low test errors in both asymmetric and symmetric noises.

\subsubsection{{Results with Real-World Noise}}
Tables \ref{table:webvision} and \ref{table:food101n} summarize the results on Webvision 1.0 and FOOD-101N. \algname{} maintained its dominance over multiple state-of-the art methods for \emph{real-world} label noise as well. It improved the top-1 validation error by $4.44pp$--$8.90pp$ and $1.09pp$--$3.85pp$ in Webvision 1.0 and FOOD-101N, respectively. The lowest error of \algname{} in FOOD-101N was achieved even without extra supervision from the verification labels.

\subsection{In-depth Analysis on Selected Samples}
\label{sec:robustness_comp_details}

\subsubsection{{F1-Score of Selected Clean Samples}}

\begin{figure}[!t]
\vspace*{-0.2cm}
\includegraphics[width=8.5cm]{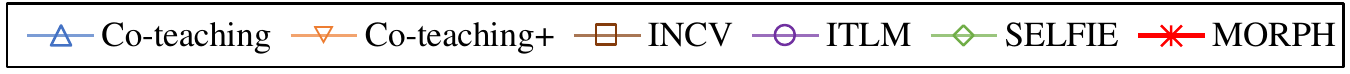}
\vspace*{-0.7cm}
\begin{center}
\includegraphics[height=30.9mm]{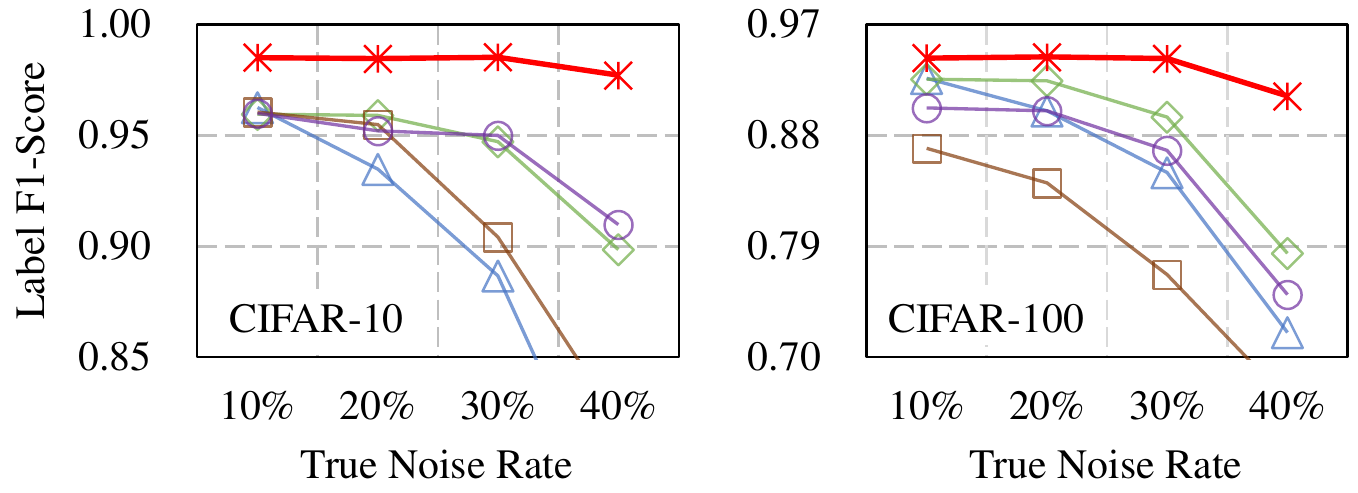}
\end{center}
\vspace*{-0.05cm}
\hspace*{1.3cm} \small{(a) Label F1-Score on Asymmetric Noise. }
\vspace*{-0.15cm}
\begin{center}
\includegraphics[height=30.9mm]{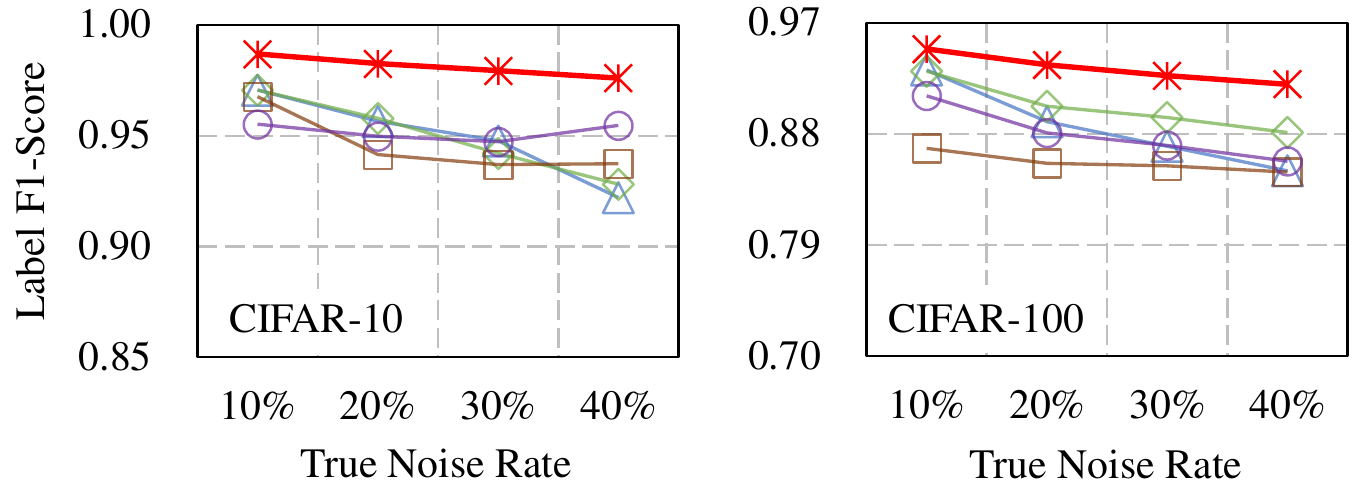}
\end{center}
\vspace*{-0.0cm}
\hspace*{1.4cm} \small{(b) Label F1-Score on Symmetric Noise. }
\vspace*{-0.25cm}
\caption{Label F1-Scores on two CIFARs data using WideResNet with varying noise rates.}
\label{fig:f_score}
\vspace*{-0.15cm}
\end{figure}

The superior robustness of \algname{} is attributed to its high \emph{label recall}\,(LR) and \emph{precision}\,(LP), which are calculated by replacing $\mathcal{M}$ for MR and MP in Eq.\,\eqref{eq:memorization_metrics} with the set of selected clean samples\,\cite{han2018co}. LR and LP are the performance indicators that respectively represent the quantity and quality of the samples selected as true-labeled ones\,\cite{han2018co}. Hence, we compare \algname{} with the other five sample selection methods in terms of the \emph{label F1-score} $= (2 \cdot LP \cdot LR)/(LP + LR)$.

\smallskip 
To evaluate the label F1-score, \algname{} used its final maximal safe set, \emph{Co-teaching(+)} and \emph{SELFIE} used the samples selected during their last epoch, and \emph{INCV} and \emph{ITLM} used the samples selected for their final training round. Figure \ref{fig:f_score} shows their label F1-scores. Only \algname{} achieved consistently high label F1-scores of over $0.91$ in all the cases. This result corroborates that \algname{} identifies true-labeled samples with high recall and precision regardless of the noise type and rate.

\subsubsection{{Evolution of the Maximal Safe Set}}
\label{sec:ablation_development}

\begin{figure}[!t]
\begin{center}
\includegraphics[width=8.5cm]{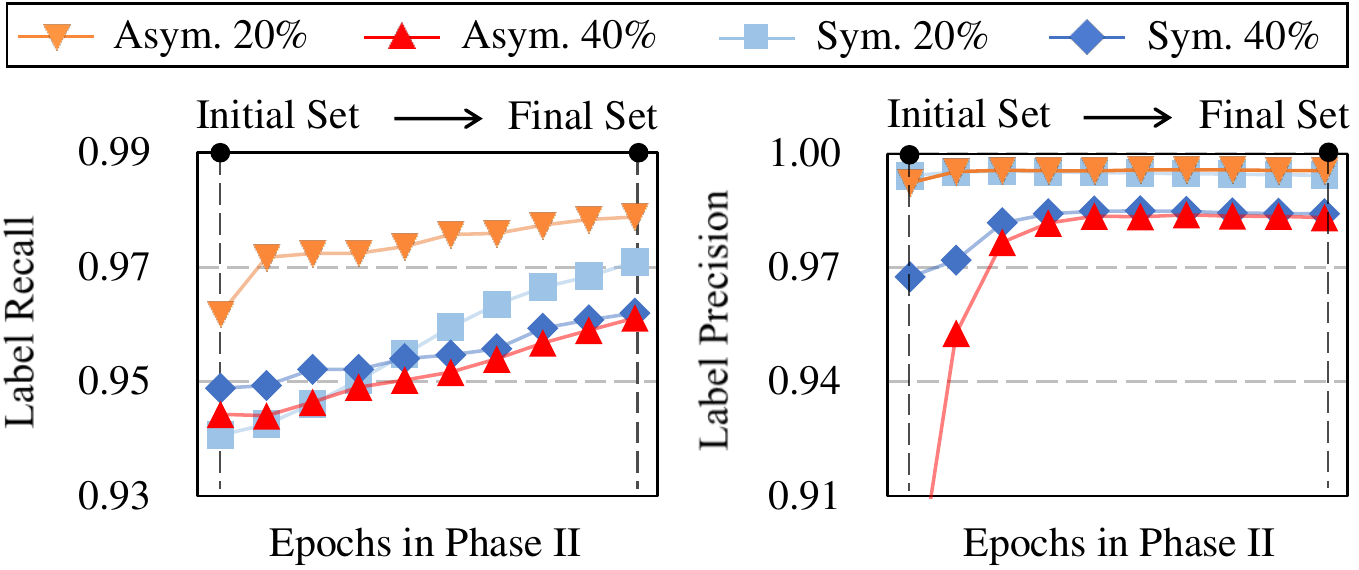}
\end{center}
\vspace*{-0.1cm}
\hspace*{1.4cm} \small{(a) Label Recall.} \hspace*{2.25cm} \small{(b) Label Precision. }
\vspace*{-0.25cm}
\caption{Evolution of the maximal safe set in Phase II using WideResNet on CIFAR-10 with two synthetic noises.}
\label{fig:evolution}
\vspace*{-0.3cm}
\end{figure}

Figure \ref{fig:evolution} shows the LR and LP values on the maximal safe set obtained at each epoch since Phase II begins. 
At the beginning of Phase II, both LR and LP already exhibited fairly high values because the initial set is derived from the samples memorized at the transition point, which is the best compromise between MR and MP. 
Moreover, as the training in Phase II progressed, LR increased gradually by adding more hard true-labeled ones previously indistinguishable\,(i.e.,  $\mathcal{C}_{new}$), and LP also increased gradually by filtering out false-labeled samples incorrectly included\,(i.e.,  $\mathcal{R}_{new}$).
Notably, their improvement was consistently observed regardless of the noise type and rate, thereby achieving remarkably high LR and LP at the end. The high F1-score of \algname{} in Figure \ref{fig:f_score} is well supported by this evolution of the maximal safe set. 

\begin{table}
\centering
\caption{Best test errors\,(\%) of {MORPH} with {early} or {late} transition based on the best point, where $\alpha$ is added to the estimated noise rate in Eq.\,(4) to force early or late transition. CIFARs with two synthetic noises of $40\%$ were used.}\label{table:transition}
\vspace*{-0.25cm}
\begin{tabular}{L{3.0cm} |X{0.65cm}X{0.65cm}|X{0.65cm}|X{0.65cm}X{0.65cm}}\toprule
Transition Point\!\! &  \multicolumn{2}{c|}{\!\!Early\!\!} & \!\!\!\!\!\!\!\!Best\!\!\!\!\!\!\!\! & \multicolumn{2}{|c}{\!\!Late\!\!} \\\midrule
A value $\alpha$ &  {\!\!\!+10\%\!\!\!} &  {\!\!+5\%\!\!} & {\!\!+0\%\!\!}  &{\!\!-5\%\!\!} & {\!\!-10\%\!\!} \\\midrule
CIFAR-10 \,\,(Sym.)\!\!\!\!    & \!\!$8.71$\!\! & \!\!$8.37$\!\! & \!\!$\textbf{8.01}$\!\! & \!\!$9.39$\!\! & \!\!$9.75$\!\! \\
CIFAR-10 \,\,(Asym.)\!\!\!\!   & \!\!$7.53$\!\! & \!\!$\textbf{6.20}$\!\! & \!\!$6.34$\!\! & \!\!$7.43$\!\! & \!\!$8.08$\!\! \\
CIFAR-100\,(Sym.)\!\!\!\!    & \!\!$29.8$\!\! & \!\!$28.5$\!\! & \!\!$\textbf{27.4}$\!\! & \!\!$28.1$\!\! & \!\!$28.5$\!\! \\
CIFAR-100\,(Asym.)\!\!\!\!   & \!\!$28.3$\!\! & \!\!$27.2$\!\! & \!\!$\textbf{26.3}$\!\! & \!\!$27.0$\!\! & \!\!$28.9$\!\! \\\bottomrule
\end{tabular}
\end{table}

\subsection{Training Efficiency Comparison}
\label{sec:efficiency_comparison}

\begin{figure}
\begin{center}
\includegraphics[width=8.5cm]{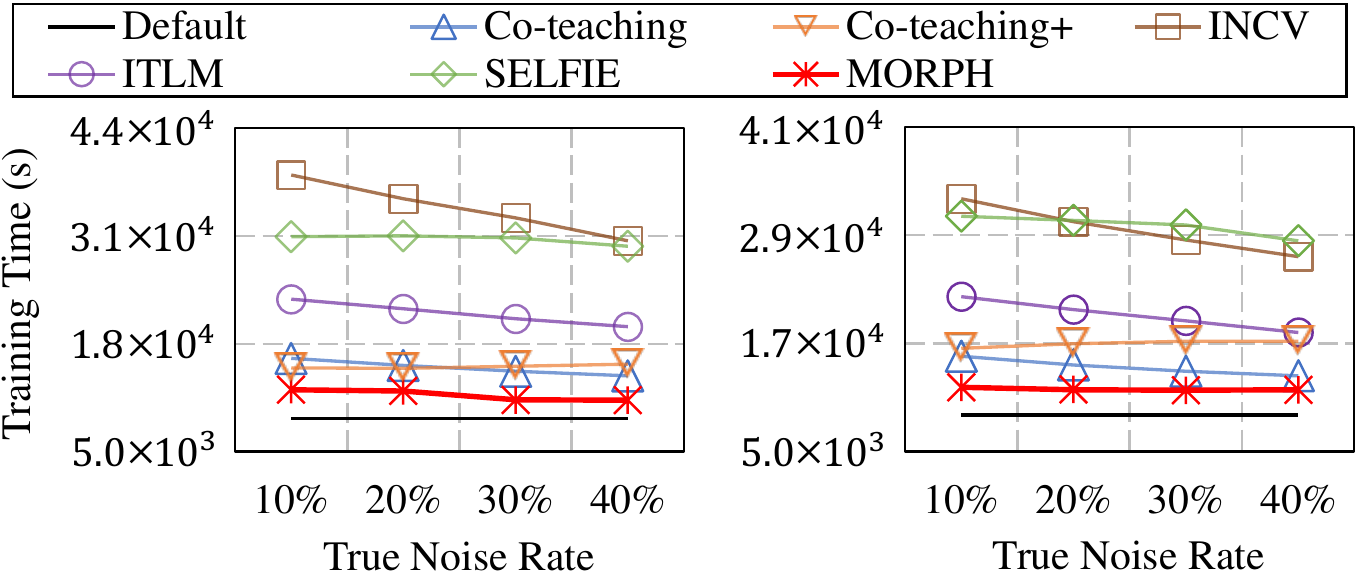}
\end{center}
\vspace*{-0.03cm}
\hspace*{1.4cm} {\small (a) CIFAR-10 Dataset.} \hspace*{1.42cm} {\small (b) CIFAR-100 Dataset.}
\vspace*{-0.3cm}
\caption{Training time on CIFAR-10 and CIFAR-100 datasets with asymmetric noise.}
\label{fig:training_time_cifar}
\vspace*{-0.3cm}
\end{figure}

Another advantage of \algname{} is its efficiency in training the DNN. Differently from the other methods that require additional DNNs or training rounds, \algname{} needs only a single training round for a single DNN. 
Figure \ref{fig:training_time_cifar} shows the training time of the six sample selection methods as well as the \emph{non}-robust method\,(``default'') on two CIFAR datasets. 
\algname{} added only $24.2$--$38.0\%$ overhead, mostly caused by the consistency regularization, compared with the non-robust method.
Among the six robust methods, \algname{} was the fastest in each case because of its relatively cheap additional costs.
Overall, \algname{} was $1.13$--$3.08$ times faster than the other methods. The difference between the training time of the robust baselines tended to be determined by the total number of training rounds, which is represented by one of their hyperparameters. 
See Table \ref{table:synthetic_training_time_table} in Appendix D.2 for the result of all datasets.  

\subsection{Optimality of the Best Transition Point}
\label{sec:optimality}
It is of interest to verify the optimality of the best transition point. Hence, we investigated the performance change of \algname{} when forcing early or late transition. 
Table \ref{table:transition} shows the test errors of \algname{} with early or late transition based on the estimated best transition point. 
The best test error was generally achieved at the estimated best point other than those around the best point. The more deviated a transition point from the best point, the higher the test error. Overall, this empirical result confirms the optimality of the best transition point estimated by \algname{}.

\begin{figure}[t]
\begin{center}
\includegraphics[width=8.4cm]{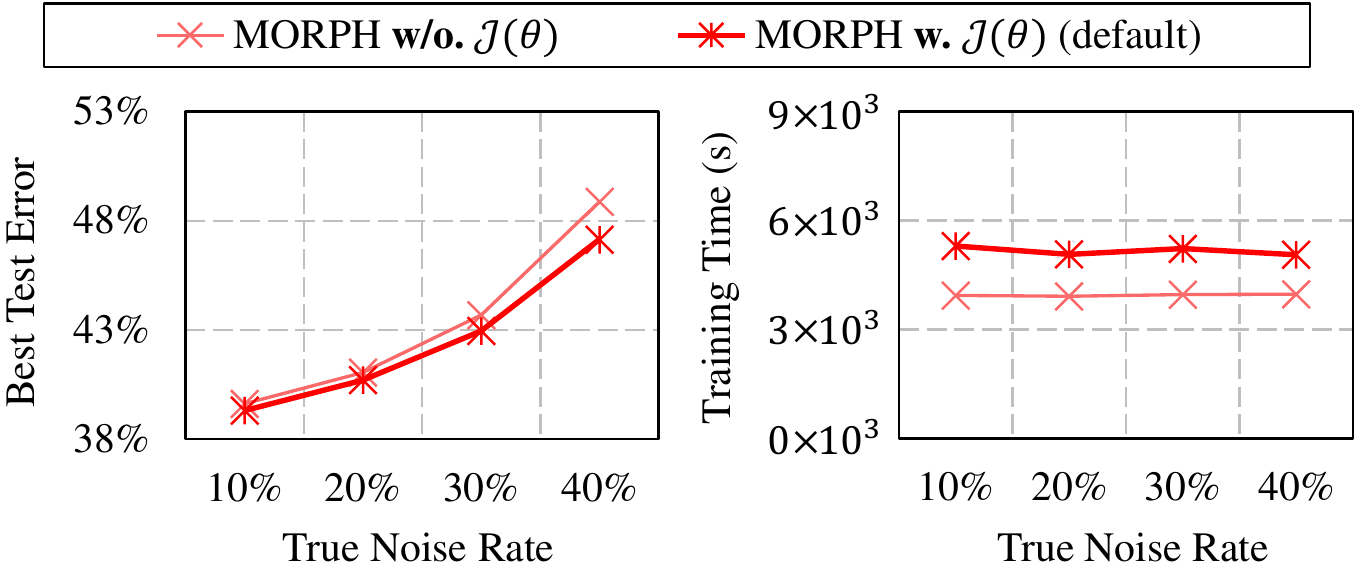}
\end{center}
\vspace*{-0.07cm}
\hspace*{1.2cm} {\small (a) Best Test Error.} \hspace*{2.2cm} {\small (b) Training Time.}
\vspace*{-0.3cm}
\caption{Effect of the consistency loss $\mathcal{J}(\theta)$ on Tiny-ImageNet with asymmetric noise.}
\label{fig:ablation_regularization}
\vspace*{-0.1cm}
\end{figure}

\subsection{Effect of Consistency Regularization}
\label{sec:consistency}

Figure \ref{fig:ablation_regularization} shows the effect of the consistency regularization. Interestingly, as shown in Figure \ref{fig:ablation_regularization}(a), the test error of \algname{} was further improved by adding the consistency loss in Eq.\ \eqref{eq:modified_safe_update}. The higher the noise rate, the greater the benefit, because the overfitting issue caused by a smaller size of the initial maximal safe set is mitigated.
Regarding the training time, as shown in Figure \ref{fig:ablation_regularization}(b), the regularization slightly slowed down the training speed owing to the extra inference steps. Because \algname{} achieved lower test error than the other robust methods even \emph{without} the regularization\,(see Appendix \ref{sec:table_error_synthetic} for details), practitioners may skip employing the regularization if their time budgets are restricted.

\section{Conclusion}

We proposed a novel \emph{self-transitional} learning scheme called \textbf{\algname{}} for noisy training data. The first phase exploits all training samples and estimates the optimal transition point. The second phase completes the rest of the training process using only a maximal safe set with high label recall and precision. \algname{} can be easily applied to many real-world cases because it requires neither a true noise rate nor a clean validation set. Through extensive experiments using various real-world and simulated noisy datasets, we verified that \algname{} consistently exhibited significant improvement in both robustness and efficiency compared with state-of-the-art methods. Overall, we believe that the division of the training process into two phases unveils a new approach to robust training and can inspire subsequent studies.
\looseness=-1

\vspace*{-0.2cm}


\section*{Acknowledgement}
This work was supported by Institute of Information \& Communications Technology Planning \& Evaluation\,(IITP) grant funded by the Korea government\,(MSIT) (No. 2020-0-00862, DB4DL: High-Usability and Performance In-Memory Distributed DBMS for Deep Learning).


\begin{thebibliography}{46}


\ifx \showCODEN    \undefined \def \showCODEN     #1{\unskip}     \fi
\ifx \showDOI      \undefined \def \showDOI       #1{#1}\fi
\ifx \showISBNx    \undefined \def \showISBNx     #1{\unskip}     \fi
\ifx \showISBNxiii \undefined \def \showISBNxiii  #1{\unskip}     \fi
\ifx \showISSN     \undefined \def \showISSN      #1{\unskip}     \fi
\ifx \showLCCN     \undefined \def \showLCCN      #1{\unskip}     \fi
\ifx \shownote     \undefined \def \shownote      #1{#1}          \fi
\ifx \showarticletitle \undefined \def \showarticletitle #1{#1}   \fi
\ifx \showURL      \undefined \def \showURL       {\relax}        \fi
\providecommand\bibfield[2]{#2}
\providecommand\bibinfo[2]{#2}
\providecommand\natexlab[1]{#1}
\providecommand\showeprint[2][]{arXiv:#2}

\bibitem[\protect\citeauthoryear{Arazo, Ortego, Albert, O’Connor, and
  Mcguinness}{Arazo et~al\mbox{.}}{2019}]%
        {arazo2019unsupervised}
\bibfield{author}{\bibinfo{person}{Eric Arazo}, \bibinfo{person}{Diego Ortego},
  \bibinfo{person}{Paul Albert}, \bibinfo{person}{Noel O’Connor}, {and}
  \bibinfo{person}{Kevin Mcguinness}.} \bibinfo{year}{2019}\natexlab{}.
\newblock \showarticletitle{Unsupervised label noise modeling and loss
  correction}. In \bibinfo{booktitle}{\emph{ICML}}. \bibinfo{pages}{312--321}.
\newblock


\bibitem[\protect\citeauthoryear{Arpit, Jastrzebski, Ballas, Krueger, Bengio,
  Kanwal, Maharaj, Fischer, Courville, Bengio, et~al\mbox{.}}{Arpit
  et~al\mbox{.}}{2017}]%
        {arpit2017closer}
\bibfield{author}{\bibinfo{person}{Devansh Arpit}, \bibinfo{person}{Stanislaw
  Jastrzebski}, \bibinfo{person}{Nicolas Ballas}, \bibinfo{person}{David
  Krueger}, \bibinfo{person}{Emmanuel Bengio}, \bibinfo{person}{Maxinder~S
  Kanwal}, \bibinfo{person}{Tegan Maharaj}, \bibinfo{person}{Asja Fischer},
  \bibinfo{person}{Aaron Courville}, \bibinfo{person}{Yoshua Bengio},
  {et~al\mbox{.}}} \bibinfo{year}{2017}\natexlab{}.
\newblock \showarticletitle{A closer look at memorization in deep networks}. In
  \bibinfo{booktitle}{\emph{ICML}}. \bibinfo{pages}{233--242}.
\newblock


\bibitem[\protect\citeauthoryear{Berthelot, Carlini, Goodfellow, Papernot,
  Oliver, and Raffel}{Berthelot et~al\mbox{.}}{2019}]%
        {berthelot2019mixmatch}
\bibfield{author}{\bibinfo{person}{David Berthelot}, \bibinfo{person}{Nicholas
  Carlini}, \bibinfo{person}{Ian Goodfellow}, \bibinfo{person}{Nicolas
  Papernot}, \bibinfo{person}{Avital Oliver}, {and} \bibinfo{person}{Colin~A
  Raffel}.} \bibinfo{year}{2019}\natexlab{}.
\newblock \showarticletitle{Mix{M}atch: {A} holistic approach to
  semi-supervised learning}. In \bibinfo{booktitle}{\emph{NeurIPS}}.
  \bibinfo{pages}{5049--5059}.
\newblock


\bibitem[\protect\citeauthoryear{Chang, Learned-Miller, and McCallum}{Chang
  et~al\mbox{.}}{2017}]%
        {chang2017active}
\bibfield{author}{\bibinfo{person}{Haw-Shiuan Chang}, \bibinfo{person}{Erik
  Learned-Miller}, {and} \bibinfo{person}{Andrew McCallum}.}
  \bibinfo{year}{2017}\natexlab{}.
\newblock \showarticletitle{Active {B}ias: Training more accurate neural
  networks by emphasizing high variance samples}. In
  \bibinfo{booktitle}{\emph{NeurIPS}}. \bibinfo{pages}{1002--1012}.
\newblock


\bibitem[\protect\citeauthoryear{Chen, Liao, Chen, and Zhang}{Chen
  et~al\mbox{.}}{2019}]%
        {chen2019understanding}
\bibfield{author}{\bibinfo{person}{Pengfei Chen}, \bibinfo{person}{Ben~Ben
  Liao}, \bibinfo{person}{Guangyong Chen}, {and} \bibinfo{person}{Shengyu
  Zhang}.} \bibinfo{year}{2019}\natexlab{}.
\newblock \showarticletitle{Understanding and utilizing deep neural networks
  trained with noisy labels}. In \bibinfo{booktitle}{\emph{ICML}}.
  \bibinfo{pages}{1062--1070}.
\newblock


\bibitem[\protect\citeauthoryear{Deng, Dong, Socher, Li, Li, and Fei-Fei}{Deng
  et~al\mbox{.}}{2009}]%
        {deng2009imagenet}
\bibfield{author}{\bibinfo{person}{Jia Deng}, \bibinfo{person}{Wei Dong},
  \bibinfo{person}{Richard Socher}, \bibinfo{person}{Li-Jia Li},
  \bibinfo{person}{Kai Li}, {and} \bibinfo{person}{Li Fei-Fei}.}
  \bibinfo{year}{2009}\natexlab{}.
\newblock \showarticletitle{{ImageNet}: A large-scale hierarchical image
  database}. In \bibinfo{booktitle}{\emph{CVPR}}. \bibinfo{pages}{248--255}.
\newblock


\bibitem[\protect\citeauthoryear{Fr{\'e}nay and Verleysen}{Fr{\'e}nay and
  Verleysen}{2013}]%
        {frenay2013classification}
\bibfield{author}{\bibinfo{person}{Beno{\^\i}t Fr{\'e}nay} {and}
  \bibinfo{person}{Michel Verleysen}.} \bibinfo{year}{2013}\natexlab{}.
\newblock \showarticletitle{Classification in the presence of label noise: {A}
  Survey}.
\newblock \bibinfo{journal}{\emph{IEEE Transactions on Neural Networks and
  Learning Systems}} \bibinfo{volume}{25}, \bibinfo{number}{5}
  (\bibinfo{year}{2013}), \bibinfo{pages}{845--869}.
\newblock


\bibitem[\protect\citeauthoryear{Han, Niu, Yu, Yao, Xu, Tsang, and
  Sugiyama}{Han et~al\mbox{.}}{2020}]%
        {han2020sigua}
\bibfield{author}{\bibinfo{person}{Bo Han}, \bibinfo{person}{Gang Niu},
  \bibinfo{person}{Xingrui Yu}, \bibinfo{person}{Quanming Yao},
  \bibinfo{person}{Miao Xu}, \bibinfo{person}{Ivor Tsang}, {and}
  \bibinfo{person}{Masashi Sugiyama}.} \bibinfo{year}{2020}\natexlab{}.
\newblock \showarticletitle{{SIGUA}: {F}orgetting may make learning with noisy
  labels more robust}. In \bibinfo{booktitle}{\emph{ICML}}.
  \bibinfo{pages}{4006--4016}.
\newblock


\bibitem[\protect\citeauthoryear{Han, Yao, Yu, Niu, Xu, Hu, Tsang, and
  Sugiyama}{Han et~al\mbox{.}}{2018}]%
        {han2018co}
\bibfield{author}{\bibinfo{person}{Bo Han}, \bibinfo{person}{Quanming Yao},
  \bibinfo{person}{Xingrui Yu}, \bibinfo{person}{Gang Niu},
  \bibinfo{person}{Miao Xu}, \bibinfo{person}{Weihua Hu}, \bibinfo{person}{Ivor
  Tsang}, {and} \bibinfo{person}{Masashi Sugiyama}.}
  \bibinfo{year}{2018}\natexlab{}.
\newblock \showarticletitle{Co-teaching: Robust training of deep neural
  networks with extremely noisy labels}. In
  \bibinfo{booktitle}{\emph{NeurIPS}}. \bibinfo{pages}{8536--8546}.
\newblock


\bibitem[\protect\citeauthoryear{Hinton, Vinyals, and Dean}{Hinton
  et~al\mbox{.}}{2015}]%
        {hinton2015distilling}
\bibfield{author}{\bibinfo{person}{Geoffrey Hinton}, \bibinfo{person}{Oriol
  Vinyals}, {and} \bibinfo{person}{Jeff Dean}.}
  \bibinfo{year}{2015}\natexlab{}.
\newblock \showarticletitle{Distilling the knowledge in a neural network}.
\newblock \bibinfo{journal}{\emph{CoRR}} (\bibinfo{year}{2015}).
\newblock


\bibitem[\protect\citeauthoryear{Huang, Qu, Jia, and Zhao}{Huang
  et~al\mbox{.}}{2019}]%
        {huang2019o2u}
\bibfield{author}{\bibinfo{person}{Jinchi Huang}, \bibinfo{person}{Lie Qu},
  \bibinfo{person}{Rongfei Jia}, {and} \bibinfo{person}{Binqiang Zhao}.}
  \bibinfo{year}{2019}\natexlab{}.
\newblock \showarticletitle{{O2U-Net}: {A} simple noisy label detection
  approach for deep neural networks}. In \bibinfo{booktitle}{\emph{ICCV}}.
  \bibinfo{pages}{3326--3334}.
\newblock


\bibitem[\protect\citeauthoryear{Jiang, Zhou, Leung, Li, and Fei-Fei}{Jiang
  et~al\mbox{.}}{2018}]%
        {jiang2017mentornet}
\bibfield{author}{\bibinfo{person}{Lu Jiang}, \bibinfo{person}{Zhengyuan Zhou},
  \bibinfo{person}{Thomas Leung}, \bibinfo{person}{Li-Jia Li}, {and}
  \bibinfo{person}{Li Fei-Fei}.} \bibinfo{year}{2018}\natexlab{}.
\newblock \showarticletitle{Mentor{N}et: Learning data-driven curriculum for
  very deep neural networks on corrupted labels}. In
  \bibinfo{booktitle}{\emph{ICML}}. \bibinfo{pages}{2309--2318}.
\newblock


\bibitem[\protect\citeauthoryear{Kamani, Farhang, Mahdavi, and Wang}{Kamani
  et~al\mbox{.}}{2020}]%
        {kamani2020targeted}
\bibfield{author}{\bibinfo{person}{Mohammad~Mahdi Kamani},
  \bibinfo{person}{Sadegh Farhang}, \bibinfo{person}{Mehrdad Mahdavi}, {and}
  \bibinfo{person}{James~Z Wang}.} \bibinfo{year}{2020}\natexlab{}.
\newblock \showarticletitle{Targeted data-driven regularization for
  out-of-distribution generalization}. In \bibinfo{booktitle}{\emph{KDD}}.
  \bibinfo{pages}{882--891}.
\newblock


\bibitem[\protect\citeauthoryear{Krizhevsky, Nair, and Hinton}{Krizhevsky
  et~al\mbox{.}}{2014}]%
        {krizhevsky2014cifar}
\bibfield{author}{\bibinfo{person}{Alex Krizhevsky}, \bibinfo{person}{Vinod
  Nair}, {and} \bibinfo{person}{Geoffrey Hinton}.}
  \bibinfo{year}{2014}\natexlab{}.
\newblock \bibinfo{title}{{CIFAR-10 and CIFAR-100} datasets}.
\newblock
\newblock
\newblock
\shownote{\url{https://www.cs.toronto.edu/~kriz/cifar.html}.}


\bibitem[\protect\citeauthoryear{Krizhevsky, Sutskever, and Hinton}{Krizhevsky
  et~al\mbox{.}}{2012}]%
        {krizhevsky2012imagenet}
\bibfield{author}{\bibinfo{person}{Alex Krizhevsky}, \bibinfo{person}{Ilya
  Sutskever}, {and} \bibinfo{person}{Geoffrey~E Hinton}.}
  \bibinfo{year}{2012}\natexlab{}.
\newblock \showarticletitle{Image{N}et classification with deep convolutional
  neural networks}. In \bibinfo{booktitle}{\emph{NeurIPS}}.
  \bibinfo{pages}{1097--1105}.
\newblock


\bibitem[\protect\citeauthoryear{Laine and Aila}{Laine and Aila}{2017}]%
        {laine2017temporal}
\bibfield{author}{\bibinfo{person}{Samuli Laine} {and} \bibinfo{person}{Tim
  Aila}.} \bibinfo{year}{2017}\natexlab{}.
\newblock \showarticletitle{Temporal ensembling for semi-supervised learning}.
  In \bibinfo{booktitle}{\emph{ICLR}}.
\newblock


\bibitem[\protect\citeauthoryear{Lee, He, Zhang, and Yang}{Lee
  et~al\mbox{.}}{2018}]%
        {lee2018cleannet}
\bibfield{author}{\bibinfo{person}{Kuang-Huei Lee}, \bibinfo{person}{Xiaodong
  He}, \bibinfo{person}{Lei Zhang}, {and} \bibinfo{person}{Linjun Yang}.}
  \bibinfo{year}{2018}\natexlab{}.
\newblock \showarticletitle{Clean{N}et: Transfer learning for scalable image
  classifier training with label noise}. In \bibinfo{booktitle}{\emph{CVPR}}.
  \bibinfo{pages}{5447--5456}.
\newblock


\bibitem[\protect\citeauthoryear{Li, Socher, and Hoi}{Li
  et~al\mbox{.}}{2020b}]%
        {li2020dividemix}
\bibfield{author}{\bibinfo{person}{Junnan Li}, \bibinfo{person}{Richard
  Socher}, {and} \bibinfo{person}{Steven~CH Hoi}.}
  \bibinfo{year}{2020}\natexlab{b}.
\newblock \showarticletitle{Divide{M}ix: Learning with noisy labels as
  semi-supervised learning}. In \bibinfo{booktitle}{\emph{ICLR}}.
\newblock


\bibitem[\protect\citeauthoryear{Li, Soltanolkotabi, and Oymak}{Li
  et~al\mbox{.}}{2020c}]%
        {li2020gradient}
\bibfield{author}{\bibinfo{person}{Mingchen Li}, \bibinfo{person}{Mahdi
  Soltanolkotabi}, {and} \bibinfo{person}{Samet Oymak}.}
  \bibinfo{year}{2020}\natexlab{c}.
\newblock \showarticletitle{Gradient descent with early stopping is provably
  robust to label noise for overparameterized neural networks}. In
  \bibinfo{booktitle}{\emph{AISTATS}}. \bibinfo{pages}{4313--4324}.
\newblock


\bibitem[\protect\citeauthoryear{Li, Peng, Cao, Du, Xing, Qiao, and Peng}{Li
  et~al\mbox{.}}{2020a}]%
        {li2020product}
\bibfield{author}{\bibinfo{person}{Qing Li}, \bibinfo{person}{Xiaojiang Peng},
  \bibinfo{person}{Liangliang Cao}, \bibinfo{person}{Wenbin Du},
  \bibinfo{person}{Hao Xing}, \bibinfo{person}{Yu Qiao}, {and}
  \bibinfo{person}{Qiang Peng}.} \bibinfo{year}{2020}\natexlab{a}.
\newblock \showarticletitle{Product image recognition with guidance learning
  and noisy supervision}.
\newblock \bibinfo{journal}{\emph{Computer Vision and Image Understanding}}
  (\bibinfo{year}{2020}), \bibinfo{pages}{102963}.
\newblock


\bibitem[\protect\citeauthoryear{Li, Wang, Li, Agustsson, and Van~Gool}{Li
  et~al\mbox{.}}{2017}]%
        {li2017webvision}
\bibfield{author}{\bibinfo{person}{Wen Li}, \bibinfo{person}{Limin Wang},
  \bibinfo{person}{Wei Li}, \bibinfo{person}{Eirikur Agustsson}, {and}
  \bibinfo{person}{Luc Van~Gool}.} \bibinfo{year}{2017}\natexlab{}.
\newblock \showarticletitle{Webvision database: Visual learning and
  understanding from web data}.
\newblock \bibinfo{journal}{\emph{arXiv preprint arXiv:1708.02862}}
  (\bibinfo{year}{2017}).
\newblock


\bibitem[\protect\citeauthoryear{Liu, Niles-Weed, Razavian, and
  Fernandez-Granda}{Liu et~al\mbox{.}}{2020}]%
        {liu2020early}
\bibfield{author}{\bibinfo{person}{Sheng Liu}, \bibinfo{person}{Jonathan
  Niles-Weed}, \bibinfo{person}{Narges Razavian}, {and} \bibinfo{person}{Carlos
  Fernandez-Granda}.} \bibinfo{year}{2020}\natexlab{}.
\newblock \showarticletitle{Early-learning regularization prevents memorization
  of noisy labels}. In \bibinfo{booktitle}{\emph{NeurIPS}}.
\newblock


\bibitem[\protect\citeauthoryear{Ma, Wang, Houle, Zhou, Erfani, Xia,
  Wijewickrema, and Bailey}{Ma et~al\mbox{.}}{2018}]%
        {ma2018dimensionality}
\bibfield{author}{\bibinfo{person}{Xingjun Ma}, \bibinfo{person}{Yisen Wang},
  \bibinfo{person}{Michael~E Houle}, \bibinfo{person}{Shuo Zhou},
  \bibinfo{person}{Sarah~M Erfani}, \bibinfo{person}{Shu-Tao Xia},
  \bibinfo{person}{Sudanthi Wijewickrema}, {and} \bibinfo{person}{James
  Bailey}.} \bibinfo{year}{2018}\natexlab{}.
\newblock \showarticletitle{Dimensionality-driven learning with noisy labels}.
  In \bibinfo{booktitle}{\emph{ICML}}. \bibinfo{pages}{3361--3370}.
\newblock


\bibitem[\protect\citeauthoryear{Ma and Leijon}{Ma and Leijon}{2011}]%
        {ma2011bayesian}
\bibfield{author}{\bibinfo{person}{Zhanyu Ma} {and} \bibinfo{person}{Arne
  Leijon}.} \bibinfo{year}{2011}\natexlab{}.
\newblock \showarticletitle{Bayesian estimation of beta mixture models with
  variational inference}.
\newblock \bibinfo{journal}{\emph{IEEE Transactions on Pattern Analysis and
  Machine Intelligence}} \bibinfo{volume}{33}, \bibinfo{number}{11}
  (\bibinfo{year}{2011}), \bibinfo{pages}{2160--2173}.
\newblock


\bibitem[\protect\citeauthoryear{Malach and Shalev-Shwartz}{Malach and
  Shalev-Shwartz}{2017}]%
        {malach2017decoupling}
\bibfield{author}{\bibinfo{person}{Eran Malach} {and} \bibinfo{person}{Shai
  Shalev-Shwartz}.} \bibinfo{year}{2017}\natexlab{}.
\newblock \showarticletitle{Decoupling ``when to update'' from ``how to
  update''}. In \bibinfo{booktitle}{\emph{NeurIPS}}. \bibinfo{pages}{960--970}.
\newblock


\bibitem[\protect\citeauthoryear{Patrini, Rozza, Menon, Nock, and Qu}{Patrini
  et~al\mbox{.}}{2017}]%
        {patrini2017making}
\bibfield{author}{\bibinfo{person}{Giorgio Patrini},
  \bibinfo{person}{Alessandro Rozza}, \bibinfo{person}{Aditya~Krishna Menon},
  \bibinfo{person}{Richard Nock}, {and} \bibinfo{person}{Lizhen Qu}.}
  \bibinfo{year}{2017}\natexlab{}.
\newblock \showarticletitle{Making deep neural networks robust to label noise:
  A loss correction approach}. In \bibinfo{booktitle}{\emph{CVPR}}.
  \bibinfo{pages}{2233--2241}.
\newblock


\bibitem[\protect\citeauthoryear{Pleiss, Zhang, Elenberg, and
  Weinberger}{Pleiss et~al\mbox{.}}{2020}]%
        {pleiss2020detecting}
\bibfield{author}{\bibinfo{person}{Geoff Pleiss}, \bibinfo{person}{Tianyi
  Zhang}, \bibinfo{person}{Ethan~R. Elenberg}, {and} \bibinfo{person}{Kilian~Q.
  Weinberger}.} \bibinfo{year}{2020}\natexlab{}.
\newblock \bibinfo{title}{Detecting noisy training data with loss curves}.
\newblock
\newblock
\urldef\tempurl%
\url{https://openreview.net/forum?id=HyenUkrtDB}
\showURL{%
\tempurl}


\bibitem[\protect\citeauthoryear{Reed, Lee, Anguelov, Szegedy, Erhan, and
  Rabinovich}{Reed et~al\mbox{.}}{2015}]%
        {reed2014training}
\bibfield{author}{\bibinfo{person}{Scott Reed}, \bibinfo{person}{Honglak Lee},
  \bibinfo{person}{Dragomir Anguelov}, \bibinfo{person}{Christian Szegedy},
  \bibinfo{person}{Dumitru Erhan}, {and} \bibinfo{person}{Andrew Rabinovich}.}
  \bibinfo{year}{2015}\natexlab{}.
\newblock \showarticletitle{Training deep neural networks on noisy labels with
  bootstrapping}. In \bibinfo{booktitle}{\emph{ICLR}}.
\newblock


\bibitem[\protect\citeauthoryear{Ren, Zeng, Yang, and Urtasun}{Ren
  et~al\mbox{.}}{2018}]%
        {ren2018learning}
\bibfield{author}{\bibinfo{person}{Mengye Ren}, \bibinfo{person}{Wenyuan Zeng},
  \bibinfo{person}{Bin Yang}, {and} \bibinfo{person}{Raquel Urtasun}.}
  \bibinfo{year}{2018}\natexlab{}.
\newblock \showarticletitle{Learning to reweight examples for robust deep
  learning}. In \bibinfo{booktitle}{\emph{ICML}}. \bibinfo{pages}{4334--4343}.
\newblock


\bibitem[\protect\citeauthoryear{Shamir and Zhang}{Shamir and Zhang}{2013}]%
        {shamir2013stochastic}
\bibfield{author}{\bibinfo{person}{Ohad Shamir} {and} \bibinfo{person}{Tong
  Zhang}.} \bibinfo{year}{2013}\natexlab{}.
\newblock \showarticletitle{Stochastic gradient descent for non-smooth
  optimization: Convergence results and optimal averaging schemes}. In
  \bibinfo{booktitle}{\emph{ICML}}. \bibinfo{pages}{71--79}.
\newblock


\bibitem[\protect\citeauthoryear{Shen and Sanghavi}{Shen and Sanghavi}{2019}]%
        {shen2019learning}
\bibfield{author}{\bibinfo{person}{Yanyao Shen} {and} \bibinfo{person}{Sujay
  Sanghavi}.} \bibinfo{year}{2019}\natexlab{}.
\newblock \showarticletitle{Learning with bad training data via iterative
  trimmed loss minimization}. In \bibinfo{booktitle}{\emph{ICML}}.
  \bibinfo{pages}{5739--5748}.
\newblock


\bibitem[\protect\citeauthoryear{Shu, Xie, Yi, Zhao, Zhou, Xu, and Meng}{Shu
  et~al\mbox{.}}{2019}]%
        {shu2019meta}
\bibfield{author}{\bibinfo{person}{Jun Shu}, \bibinfo{person}{Qi Xie},
  \bibinfo{person}{Lixuan Yi}, \bibinfo{person}{Qian Zhao},
  \bibinfo{person}{Sanping Zhou}, \bibinfo{person}{Zongben Xu}, {and}
  \bibinfo{person}{Deyu Meng}.} \bibinfo{year}{2019}\natexlab{}.
\newblock \showarticletitle{Meta-weight-net: Learning an explicit mapping for
  sample weighting}. In \bibinfo{booktitle}{\emph{NeurIPS}}.
  \bibinfo{pages}{1917--1928}.
\newblock


\bibitem[\protect\citeauthoryear{Song, Kim, and Lee}{Song
  et~al\mbox{.}}{2019}]%
        {song2019selfie}
\bibfield{author}{\bibinfo{person}{Hwanjun Song}, \bibinfo{person}{Minseok
  Kim}, {and} \bibinfo{person}{Jae-Gil Lee}.} \bibinfo{year}{2019}\natexlab{}.
\newblock \showarticletitle{{SELFIE}: Refurbishing unclean samples for robust
  deep learning}. In \bibinfo{booktitle}{\emph{ICML}}.
  \bibinfo{pages}{5907--5915}.
\newblock


\bibitem[\protect\citeauthoryear{Song, Kim, Park, and Lee}{Song
  et~al\mbox{.}}{2020a}]%
        {song2020prestopping}
\bibfield{author}{\bibinfo{person}{Hwanjun Song}, \bibinfo{person}{Minseok
  Kim}, \bibinfo{person}{Dongmin Park}, {and} \bibinfo{person}{Jae-Gil Lee}.}
  \bibinfo{year}{2020}\natexlab{a}.
\newblock \showarticletitle{How does early stopping help generalization against
  label noise?}. In \bibinfo{booktitle}{\emph{ICMLW}}.
\newblock


\bibitem[\protect\citeauthoryear{Song, Kim, Park, and Lee}{Song
  et~al\mbox{.}}{2020b}]%
        {song2020learning}
\bibfield{author}{\bibinfo{person}{Hwanjun Song}, \bibinfo{person}{Minseok
  Kim}, \bibinfo{person}{Dongmin Park}, {and} \bibinfo{person}{Jae-Gil Lee}.}
  \bibinfo{year}{2020}\natexlab{b}.
\newblock \showarticletitle{Learning from noisy labels with deep neural
  networks: A survey}.
\newblock \bibinfo{journal}{\emph{arXiv preprint arXiv:2007.08199}}
  (\bibinfo{year}{2020}).
\newblock


\bibitem[\protect\citeauthoryear{Szegedy, Ioffe, Vanhoucke, and Alemi}{Szegedy
  et~al\mbox{.}}{2017}]%
        {szegedy2017inception}
\bibfield{author}{\bibinfo{person}{Christian Szegedy}, \bibinfo{person}{Sergey
  Ioffe}, \bibinfo{person}{Vincent Vanhoucke}, {and}
  \bibinfo{person}{Alexander~A Alemi}.} \bibinfo{year}{2017}\natexlab{}.
\newblock \showarticletitle{Inception-v4, inception-resnet and the impact of
  residual connections on learning}. In \bibinfo{booktitle}{\emph{AAAI}}.
\newblock


\bibitem[\protect\citeauthoryear{Tang, Borisyuk, Malreddy, Li, Liu, and
  Kirshner}{Tang et~al\mbox{.}}{2019}]%
        {tang2019msuru}
\bibfield{author}{\bibinfo{person}{Yina Tang}, \bibinfo{person}{Fedor
  Borisyuk}, \bibinfo{person}{Siddarth Malreddy}, \bibinfo{person}{Yixuan Li},
  \bibinfo{person}{Yiqun Liu}, {and} \bibinfo{person}{Sergey Kirshner}.}
  \bibinfo{year}{2019}\natexlab{}.
\newblock \showarticletitle{{MSURU}: Large scale e-commerce image
  classification with weakly supervised search data}. In
  \bibinfo{booktitle}{\emph{KDD}}. \bibinfo{pages}{2518--2526}.
\newblock


\bibitem[\protect\citeauthoryear{Tarvainen and Valpola}{Tarvainen and
  Valpola}{2017}]%
        {tarvainen2017mean}
\bibfield{author}{\bibinfo{person}{Antti Tarvainen} {and}
  \bibinfo{person}{Harri Valpola}.} \bibinfo{year}{2017}\natexlab{}.
\newblock \showarticletitle{Mean teachers are better role models:
  Weight-averaged consistency targets improve semi-supervised deep learning
  results}. In \bibinfo{booktitle}{\emph{NeurIPS}}.
  \bibinfo{pages}{1195--1204}.
\newblock


\bibitem[\protect\citeauthoryear{Toneva, Sordoni, Combes, Trischler, Bengio,
  and Gordon}{Toneva et~al\mbox{.}}{2019}]%
        {toneva2018empirical}
\bibfield{author}{\bibinfo{person}{Mariya Toneva}, \bibinfo{person}{Alessandro
  Sordoni}, \bibinfo{person}{Remi Tachet~des Combes}, \bibinfo{person}{Adam
  Trischler}, \bibinfo{person}{Yoshua Bengio}, {and}
  \bibinfo{person}{Geoffrey~J Gordon}.} \bibinfo{year}{2019}\natexlab{}.
\newblock \showarticletitle{An empirical study of example forgetting during
  deep neural network learning}. In \bibinfo{booktitle}{\emph{ICLR}}.
\newblock


\bibitem[\protect\citeauthoryear{Wang, Liu, Ma, Bailey, Zha, Song, and
  Xia}{Wang et~al\mbox{.}}{2018}]%
        {wang2018iterative}
\bibfield{author}{\bibinfo{person}{Yisen Wang}, \bibinfo{person}{Weiyang Liu},
  \bibinfo{person}{Xingjun Ma}, \bibinfo{person}{James Bailey},
  \bibinfo{person}{Hongyuan Zha}, \bibinfo{person}{Le Song}, {and}
  \bibinfo{person}{Shu-Tao Xia}.} \bibinfo{year}{2018}\natexlab{}.
\newblock \showarticletitle{Iterative learning with open-set noisy labels}. In
  \bibinfo{booktitle}{\emph{CVPR}}. \bibinfo{pages}{8688--8696}.
\newblock


\bibitem[\protect\citeauthoryear{Xiao, Xiao, and Eckert}{Xiao
  et~al\mbox{.}}{2012}]%
        {xiao2012adversarial}
\bibfield{author}{\bibinfo{person}{Han Xiao}, \bibinfo{person}{Huang Xiao},
  {and} \bibinfo{person}{Claudia Eckert}.} \bibinfo{year}{2012}\natexlab{}.
\newblock \showarticletitle{Adversarial label flips attack on support vector
  machines}. In \bibinfo{booktitle}{\emph{ECAI}}. \bibinfo{pages}{870--875}.
\newblock


\bibitem[\protect\citeauthoryear{Yu, Han, Yao, Niu, Tsang, and Sugiyama}{Yu
  et~al\mbox{.}}{2019}]%
        {yu2019does}
\bibfield{author}{\bibinfo{person}{Xingrui Yu}, \bibinfo{person}{Bo Han},
  \bibinfo{person}{Jiangchao Yao}, \bibinfo{person}{Gang Niu},
  \bibinfo{person}{Ivor Tsang}, {and} \bibinfo{person}{Masashi Sugiyama}.}
  \bibinfo{year}{2019}\natexlab{}.
\newblock \showarticletitle{How does disagreement help generalization against
  label corruption?}. In \bibinfo{booktitle}{\emph{ICML}}.
  \bibinfo{pages}{7164--7173}.
\newblock


\bibitem[\protect\citeauthoryear{Zhang, Bengio, Hardt, Mozer, and Singer}{Zhang
  et~al\mbox{.}}{2020a}]%
        {zhang2019identity}
\bibfield{author}{\bibinfo{person}{Chiyuan Zhang}, \bibinfo{person}{Samy
  Bengio}, \bibinfo{person}{Moritz Hardt}, \bibinfo{person}{Michael~C Mozer},
  {and} \bibinfo{person}{Yoram Singer}.} \bibinfo{year}{2020}\natexlab{a}.
\newblock \showarticletitle{Identity {C}risis: Memorization and generalization
  under extreme overparameterization}. In \bibinfo{booktitle}{\emph{ICLR}}.
\newblock


\bibitem[\protect\citeauthoryear{Zhang, Bengio, Hardt, Recht, and
  Vinyals}{Zhang et~al\mbox{.}}{2017}]%
        {zhang2016understanding}
\bibfield{author}{\bibinfo{person}{Chiyuan Zhang}, \bibinfo{person}{Samy
  Bengio}, \bibinfo{person}{Moritz Hardt}, \bibinfo{person}{Benjamin Recht},
  {and} \bibinfo{person}{Oriol Vinyals}.} \bibinfo{year}{2017}\natexlab{}.
\newblock \showarticletitle{Understanding deep learning requires rethinking
  generalization}. In \bibinfo{booktitle}{\emph{ICLR}}.
\newblock


\bibitem[\protect\citeauthoryear{Zhang, Zhang, Odena, and Lee}{Zhang
  et~al\mbox{.}}{2020b}]%
        {zhang2019consistency}
\bibfield{author}{\bibinfo{person}{Han Zhang}, \bibinfo{person}{Zizhao Zhang},
  \bibinfo{person}{Augustus Odena}, {and} \bibinfo{person}{Honglak Lee}.}
  \bibinfo{year}{2020}\natexlab{b}.
\newblock \showarticletitle{Consistency regularization for generative
  adversarial networks}. In \bibinfo{booktitle}{\emph{ICLR}}.
\newblock


\bibitem[\protect\citeauthoryear{Zhuang, Liu, Li, Shen, and Reid}{Zhuang
  et~al\mbox{.}}{2017}]%
        {zhuang2017attend}
\bibfield{author}{\bibinfo{person}{Bohan Zhuang}, \bibinfo{person}{Lingqiao
  Liu}, \bibinfo{person}{Yao Li}, \bibinfo{person}{Chunhua Shen}, {and}
  \bibinfo{person}{Ian Reid}.} \bibinfo{year}{2017}\natexlab{}.
\newblock \showarticletitle{Attend in groups: A weakly-supervised deep learning
  framework for learning from web data}. In \bibinfo{booktitle}{\emph{CVPR}}.
  \bibinfo{pages}{1878--1887}.
\newblock


\end{thebibliography}

\clearpage
\begin{appendix}
\begin{figure}[!htb]
\begin{center}
\includegraphics[width=8cm]{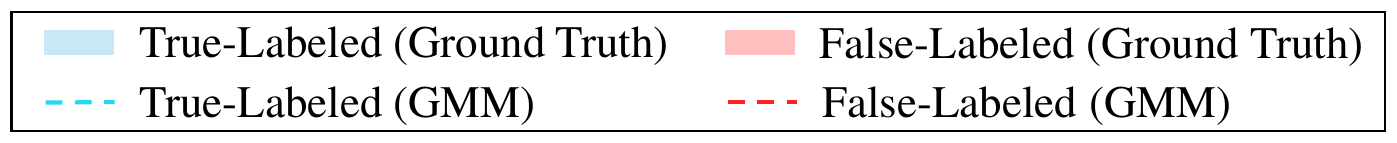}
\vspace{-0.4cm}
\end{center}
\begin{center}
\includegraphics[height=27mm]{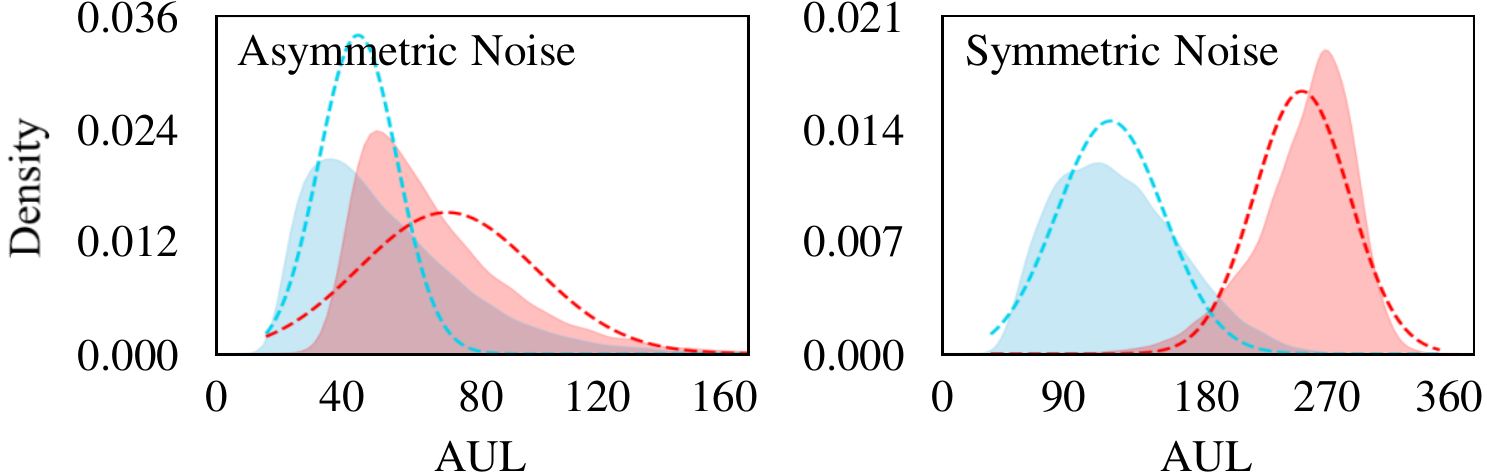}\\
\vspace*{-0.15cm}
\hspace*{0.55cm} \small{(a) CIFAR-10 Dataset.} 
\includegraphics[height=27mm]{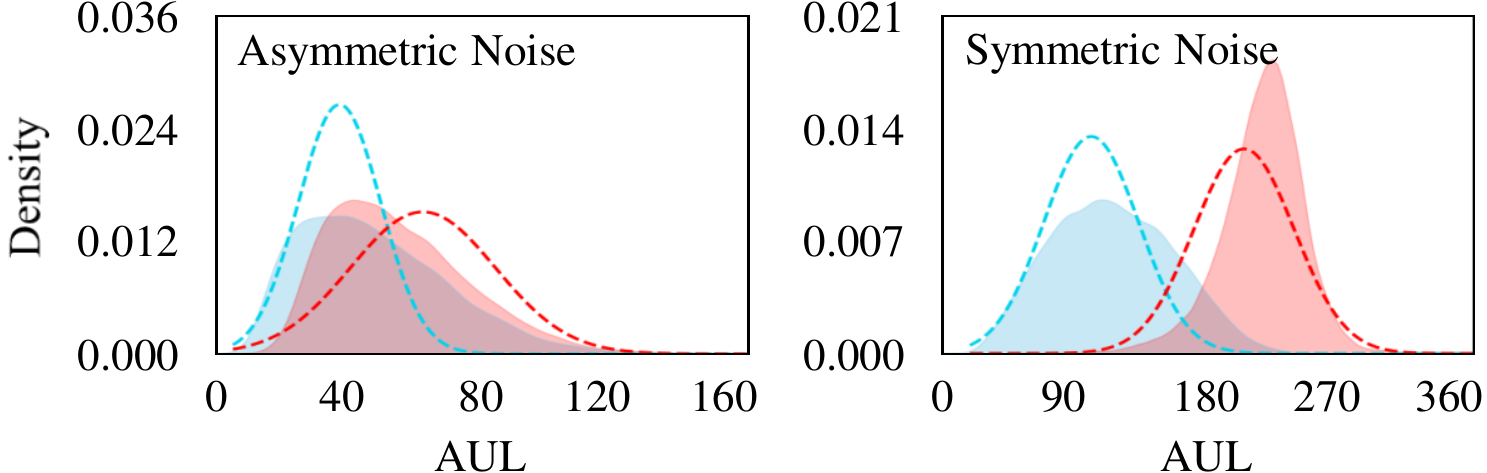}\\
\vspace*{-0.15cm}
\hspace*{0.8cm} \small{(b) CIFAR-100 Dataset. }
\end{center}
\vspace*{-0.35cm}
\caption{AUL distributions of true- and false-labeled samples using the ground-truth label and the GMM on two CIFAR datasets with two synthetic noises of $40\%$.}
\vspace*{-0.4cm}
\label{fig:gmm_analysis}
\vspace*{-0.1cm}
\end{figure}

\section{Noise Rate Estimation}
\label{sec:generalization}

In order to accurately estimate the noise rate, we compared \emph{two} widely-used methods. The first one was adopted in all experiments because it showed better performance than the second one.
\vspace*{+0.01cm}
\begin{enumerate}[label={\arabic*.}, leftmargin=9pt]
\item \textbf{Gaussian Mixture Model\,(GMM):} The first method is exploiting a one-dimensional and two-component GMM to model the loss distribution of true- and false-labeled samples\,\cite{arazo2019unsupervised, pleiss2020detecting}. Because the loss distribution tends to be bimodal, the probability of a sample being a false-labeled sample is obtained through its posterior probability. Subsequently, the noise rate is estimated by computing the expectation of the posterior probability for all the training samples.
However, in considering that the DNN eventually memorizes all the training samples, the training loss becomes less separable by the GMM as the training progresses. Thus, we computed the \emph{Area Under the Loss curve}\,(AUL)\,\cite{pleiss2020detecting}, which is the sum of the samples' training losses obtained from all previous training epochs. The main benefit of the AUL is that its distribution remains separable even after the loss signal decays in later epochs. Therefore, as shown in Figure \ref{fig:gmm_analysis}, the loss distributions of true- and false-labeled samples are modeled by fitting the GMM to the AULs of all the training samples, and the noise rate at time $t$ is estimated by
\begin{equation}
\label{eq:area_under_loss}
\begin{gathered}
\hat{\tau} = \mathbb{E}_{(x,\tilde{y})\in\tilde{\mathcal{D}}}[p\big(g|AUL_{t}(x,\tilde{y})\big)],\\ 
\text{where}~ \text{AUL}_{t}(x,\tilde{y})= \sum_{i=1}^{t}f_{(x, \tilde{y})}(\Theta_t) ~{\rm and}
\end{gathered}
\end{equation}
$g$ is the Gaussian component with a larger mean\,(i.e., larger AUL).
\vspace*{+0.01cm}
\item \textbf{Cross-Validation:} The second method is estimating the noise rate by applying cross-validation on two randomly divided noisy training datasets $\tilde{\mathcal{D}_{1}}$ and $\tilde{\mathcal{D}_{2}}$\,\cite{chen2019understanding}. Under the assumptions that these two datasets share exactly the same label transition matrix, the noise rate quantifies the test accuracy of DNNs, which are trained and tested on previously mentioned noisy datasets $\tilde{\mathcal{D}_{1}}$ and $\tilde{\mathcal{D}_{2}}$, respectively. In the case of synthetic noises, the test accuracy is approximated by a quadratic function of the noise rate. Therefore, the noise rate can be estimated from the test accuracy obtained by the cross-validation. Refer to \cite{chen2019understanding} for details.
\end{enumerate}

\vspace*{0.05cm}
\section{Supplementary Evaluation}
\vspace*{0.05cm}

\subsection{Hyperparameter Selection}
\label{sec:hyper}

\begin{figure}
\begin{center}
\includegraphics[width=8.3cm]{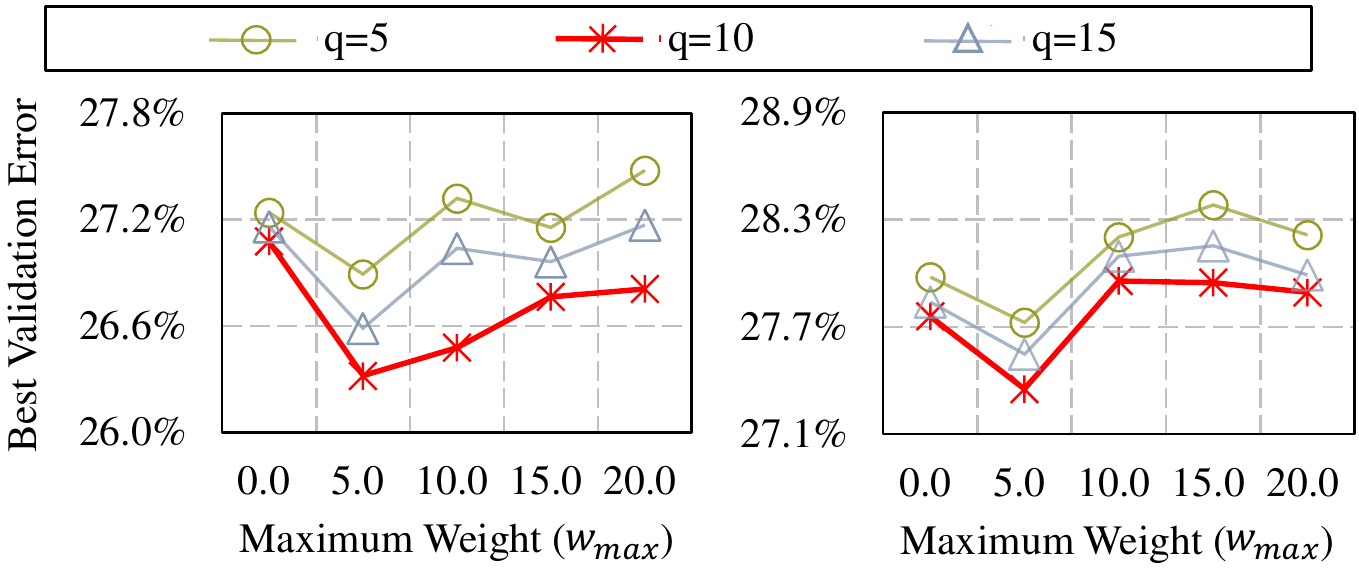}
\end{center}
\vspace*{-0.1cm}
\hspace*{1.4cm} {\small (a) Asymmetric Noise.} \hspace*{1.3cm} {\small (b) Symmetric Noise.}
\vspace*{-0.3cm}
\caption{Hyperparameter selection on the CIFAR-100 dataset with two noise types of $40\%$.}
\label{fig:grid_search}
\vspace*{-0.0cm}
\end{figure}

To ascertain the optimal values of the history length $q$ and the maximum weight $w_{max}$, we trained a WideResNet-16-8 on CIFAR-100 at a noise rate of $40\%$. Here, because no validation data exists for the CIFAR-100 dataset, we constructed a small clean validation set by randomly selecting $1,000$ images from the original training set of $50,000$ images. Then, the noise injection process was applied to only the rest $49,000$ training images.
Figure \ref{fig:grid_search} shows the validation errors of \algname{} obtained by grid search on the noisy CIFAR-100 dataset. These two hyperparameters were chosen in the grid $q \in \{5, 10, 15\}$ and $w_{max} \in \{0.0, 5.0, 10.0, 15.0\}$. 
The two hyperparameters were not that sensitive to the noise rate. Typically, the lowest validation error depending on $q$ was achieved when the value of $q$ was $10$ for both noise types. As for $w_{max}$, the validation error was observed to be the lowest when the value of $w_{max}$ was $5.0$. Therefore, we set the values of $q$ and $w_{max}$ to be $10$ and $5.0$, respectively, in all experiments.

\subsection{Comparison with More Recent Methods}
\label{sec:com_recent}

\begin{table}[t!]
\vspace*{-0.2cm}
\centering
\small
\caption{Best test errors\,(\%) of {MORPH} compared with {DM-DYR-SH} and {DivdeMix} using CIFAR-100.}\label{table:sota}
\vspace*{-0.3cm}
\begin{tabular}{L{2.3cm} |X{0.8cm}X{0.8cm}X{0.8cm}|X{0.8cm}X{0.8cm}}\toprule
Noise Type\!\! &  \multicolumn{3}{c|}{\!\!Sym. Noise\!\!} & \multicolumn{2}{c}{\!\!Asym. Noise\!\!} \\\midrule
Noise Rate &  {\!\!20\%\!\!} &  {\!\!40\%\!\!} & {\!\!70\%\!\!}  &{\!\!20\%\!\!} & {\!\!40\%\!\!} \\\toprule
\!{DM-DYR-SH}\!\!\!\!    & \!\!$28.0$\!\! & \!\!$32.2$\!\! & \!\!$46.4$\!\! & \!\!$31.1$\!\! & \!\!$46.3$\!\! \\
\!{DivdeMix}\!\!\!\!   & \!\!$25.3$\!\! & \!\!$27.6$\!\! & \!\!$\textbf{38.9}$\!\! & \!\!$25.4$\!\! & \!\!$44.1$\!\! \\
\!\textbf{{MORPH}}\!\!\!\!    & \!\!$\textbf{23.6}$\!\! & \!\!$\textbf{27.4}$\!\! & \!\!$39.5$\!\! & \!\!$\textbf{23.0}$\!\! & \!\!$\textbf{26.3}$\!\! \\\bottomrule
\end{tabular}
\vspace*{-0.2cm}
\end{table} 

\begin{table*}[!t]
\noindent
\scriptsize
\begin{center}
\parbox{.56\textwidth}{%
\parbox{9.6cm}{
\vspace*{0.1cm}
\caption{Best test errors\,(\%) of seven training methods on two \textbf{synthetic} noises of $40\%$ in Figures \ref{fig:wrn_error_pair} and \ref{fig:wrn_error_symmetric}.}
\label{table:synthetic_table}
\vspace*{-0.4cm}
}
\begin{tabular}{L{1.8cm} |X{0.9cm}X{0.9cm}X{0.9cm}|X{0.9cm}X{0.9cm}X{0.9cm}}\toprule
\!\!Noise Type &  \multicolumn{3}{c|}{Asymmetric Noise in Figure \ref{fig:wrn_error_pair}} & \multicolumn{3}{c}{Symmetric Noise in Figure \ref{fig:wrn_error_symmetric}} \\\midrule
\!Dataset &  {\!\!CIFAR-10\!\!} &  {\!\!CIFAR-100\!\!} & {\!\!\!Tiny-Image\!\!\!}  &{\!\!CIFAR-10\!\!} & {\!\!CIFAR-100\!\!} & {\!\!\!Tiny-Image\!\!\!}   \\\toprule
\!\emph{Co-teaching}\!    & $24.7$ &  $44.8$ & $61.8$ &  $12.4$ &   $30.2$ & $48.8$  \\
\!\emph{Co-teaching+}\!   & $21.3$ & $53.3$ & $66.5$ & $20.3$ & $53.7$ & $64.2$ \\
\!\emph{INCV}\!           & $22.2$ &  $47.0$ &  $63.8$ &  $9.43$ &   $30.1$ &  $49.7$  \\
\!\emph{ITLM}\!           & $12.9$ &  $41.9$ &  $63.9$  &  $8.43$ &   $29.3$ &  $49.0$ \\
\!\emph{SELFIE}\!         & $13.5$ &  $38.6$ &  $55.4$ &  $10.4$ &   $28.2$ &  $48.4$  \\\midrule
\!\textbf{\algname{} {\scriptsize w/o. $\mathcal{J}(\Theta$)}}\! & $7.25$ &  $27.1$ & $48.9$  &  $8.27$ &   $27.8$ &  $47.0$  \\
\!\textbf{\algname{} {\scriptsize w. $\mathcal{J}(\Theta$)}}\! & $\textbf{6.34}$ &  $\textbf{26.3}$ &  $\textbf{47.2}$ &  $\textbf{8.01}$ &   $\textbf{27.4}$ &  $\textbf{46.2}$  \\\bottomrule
\end{tabular}
\vspace*{0.1cm}
}%
\parbox{.55\textwidth}{%
\vspace*{-0.1cm}
\parbox{8.0cm}{
\vspace*{0.1cm}
\caption{Training time\,(sec) of seven training methods on two \textbf{synthetic} noises of $40\%$ in Figures \ref{fig:wrn_error_pair} and \ref{fig:wrn_error_symmetric}.}
\label{table:synthetic_training_time_table}
\vspace*{-0.4cm}
}
\begin{tabular}{|X{0.9cm}X{0.9cm}X{1.0cm}|X{1.0cm}X{1.0cm}X{1.0cm}}\toprule
\multicolumn{3}{|c|}{Asymmetric Noise in Figure \ref{fig:wrn_error_pair}} & \multicolumn{3}{c}{Symmetric Noise in Figure \ref{fig:wrn_error_symmetric}} \\\midrule
{\!\!CIFAR-10\!\!} &  {\!\!CIFAR-100\!\!} & {\!\!\!Tiny-Image\!\!\!}  &{\!\!CIFAR-10\!\!} & {\!\!CIFAR-100\!\!} & {\!\!\!Tiny-Image\!\!\!}   \\\toprule
14,107 &  13,386 &  60,262 &  13,171 &  11,886 &  53,660 \\
15,527 &  17,244 &  82,982 &  15,365 &  17,459 & 80,038 \\
30,376 &  26,628 &  104,597 &  28,915 &   24,668 &  99,314 \\
20,068 &  18,212 &  69,037 &  18,763 &   16,749 &  60,469 \\
29,730 &  28,352 &  115,208 &  29,436 &   27,931 &  113,380 \\\midrule
\textbf{9,646} &  \textbf{9,477} &  \textbf{39,728} &  \textbf{9,598} &   \textbf{9,488} &  \textbf{40,154} \\
{11,198} &  {11,828} &  {50,594} &  {11,331} &   12,049 &  49,412 \\\bottomrule
\end{tabular}
}
\end{center}
\vspace*{-0.5cm}
\end{table*}

\emph{DM-DYR-SH}\,\cite{arazo2019unsupervised} and \emph{DivideMix}\,\cite{li2020dividemix} are similar to \algname{} in that they also use mixture models for handling noisy labels. However, the mixture model is used for a \emph{different} purpose: \emph{DM-DYR-SH} and \emph{DivideMix} use the mixture model to identify clean samples based on the small loss trick; on the other hand, \algname{} uses the model to find the best transition point. In addition, \algname{} no longer relies on the small-loss trick, and incrementally identifies a better set of clean samples through our alternating scheme in Phase II.

Nevertheless, we compared \algname{} with \emph{DM-DYR-SH} and \emph{DivideMix} for a more in-depth analysis. The results are summarized in Table \ref{table:sota}. Only \algname{} achieved the noise type robustness; the performance of \emph{DM-DYR-SH} and \emph{DivideMix} was considerably worse in asymmetric noise because their underlying philosophy is based on the \emph{small-loss} trick. In fact, \emph{DM-DYR-SH} and \emph{DivideMix} are favored because they are equipped with an unsupervised or semi-supervised method such as \emph{Beta Mixture}\,\cite{ma2011bayesian} and \emph{MixMatch}\,\cite{berthelot2019mixmatch} to further improve the performance, while \algname{} deals with only sample selection. Please note that this superior performance of \algname{} was achieved even without adopting the unsupervised or semi-supervised method. \algname{} is expected to perform even much better when combined with such an additional technique. 

\section{Details on Experiment Setting}

\subsection{Benchmark Datasets}
\label{sec:data_decription}

For the experiment, we prepared \emph{five} synthetic or real-world noisy datasets:
CIFAR-10\footnote{\scriptsize{}\href{url}{https://www.cs.toronto.edu/\symbol{126}kriz/cifar.html}}\,($10$ classes)\,\cite{krizhevsky2014cifar} and CIFAR-100\footnotemark[3]\,($100$ classes)\,\cite{krizhevsky2014cifar}, a subset of $80$ million categorical images with $50$K training images and $10$K test images; Tiny-ImageNet\footnote{\scriptsize{}\href{url}{https://www.kaggle.com/c/tiny-imagenet}}\,($200$ classes)\,\cite{krizhevsky2012imagenet}, a subset of ImageNet with $100$K training images and $10$K test images; Webvision 1.0\footnote{\scriptsize{}\href{url}{https://data.vision.ee.ethz.ch/cvl/webvision/dataset2017.html}}\,($1,000$ classes)\,\cite{li2017webvision}, real-world noisy data of crawled images using the concepts in ImageNet with $2.4$M training images, $50$K WebVision validation images, and $50$K ImageNet ILSVRC12 validation images; FOOD-101N\footnote{\scriptsize{}\href{url}{https://kuanghuei.github.io/Food-101N}}\,($101$ classes)\,\,\cite{lee2018cleannet}, real-world noisy data of crawled food images with $310$K training images and $25$K FOOD-101 validation images. Following the previous work \cite{chen2019understanding}, we used only the first $50$ classes of the Google image subset in Webvision 1.0.

\subsection{Training Setup for Real-World Data}
\label{sec:training_configuration}

To verify the practical usability of \algname{} on real-world noisy labels, we performed a classification task on Webvision 1.0 and FOOD-101N. We followed exactly the same configurations in the previous work\,\cite{chen2019understanding, lee2018cleannet, li2020product}. For Webvision 1.0, we trained an InceptionResNet-V2 from scratch for $120$ epochs using SGD with a momentum of $0.9$ and an initial learning rate of $0.1$, which was divided by $10$ after $40$ and $80$ epochs\,(refer to \cite{chen2019understanding}). For FOOD-101N, we trained a ResNet-50 with the ImageNet pretrained weights for $60$ epochs using SGD with a momentum of $0.9$ and an initial learning rate of $0.01$, which was divided by $10$ after $30$ epochs\,(refer to \cite{lee2018cleannet}). Regardless of the dataset, we used a batch size of $64$, a dropout of $0.4$, and a weight decay of $0.001$. Random crops and horizontal flips were applied for data augmentation.

\vspace*{-0.1cm}
\subsection{Algorithm Hyperparameters}
\label{sec:compared_algorithm}
\vspace*{-0.05cm}

For reproducibility, we clarify the hyperparameter setup of all compared algorithms. Please note, if the true noise rate $\tau$ was needed as one of the hyperparameters (in \emph{Co-teaching}, \emph{Co-teaching+}, \emph{ITLM}, and \emph{SELFIE}), then it was set to be the  noise rate $\hat{\tau}$ estimated by \algname{} for fair comparison. The other hyperparameters for each method are configured favorably as follows: \looseness=-1
\begin{itemize}[leftmargin=9pt]
\item \textbf{\emph{Co-teaching(+)}:} To decrease the number of selected samples gradually at the beginning of the training, the warm-up epoch is required as a hyperparameter; it was set to be $15$, which is reported to achieve the best performance in the original paper\,\cite{han2018co}. 
\item \textbf{\emph{INCV}:} Following the original paper\,\cite{chen2019understanding}, the total number of training rounds was set to be $4$; the DNN was trained for $50$ epochs using the Adam optimizer; an initial learning rate was set to be $0.001$, which was divided by $2$ after $20$ and $30$ epochs and finally fixed to be $0.0001$ after $40$ epochs. Subsequently, all the training samples selected by \emph{INCV} were used to retrain the DNN using \emph{Co-teaching} with the same configuration.
\item \textbf{\emph{ITLM}:} Because it iterates the training process multiple times as well, the total number of training rounds was set to be $5$. As mentioned in the original paper\,\cite{shen2019learning}, the training process for the first $4$ rounds was early stopped because it may help filter out false-labeled samples. Subsequently, without early stopping, the DNN was retrained using the samples selected from the $4$th round during the last training round.
\item \textbf{\emph{SELFIE}:} Four hyperparameters are required for \emph{SELFIE}. The warm-up epoch for sample selection was set to be $15$ similar to \emph{Co-teaching}. The uncertainty threshold and the history length for loss correction were set to be $0.05$ and $15$, respectively, which are the best values obtained from a grid search in the original paper\,\cite{song2019selfie}. The training process was restarted twice according to the authors' recommendation.
\end{itemize}

\vspace*{-0.1cm}
\section{Complete Experiment Results}

\subsection{Test Error with Synthetic Noises}
\label{sec:table_error_synthetic}

Table \ref{table:synthetic_table} shows the \emph{test error} of seven training methods using WideResNet-16-8 on \emph{simulated} noisy datasets with $40\%$ noise rate. Two variants of \algname{} depending on the existence of the consistency regularization were included for comparison.

\vspace*{-0.1cm}
\subsection{Training Time with Synthetic Noises}
\label{sec:table_time_synthetic}

Table \ref{table:synthetic_training_time_table} shows the \emph{training time} of seven training methods using WideResNet-16-8 on \emph{simulated} noisy datasets with $40\%$ noise rate.

\end{appendix}

\end{document}